\documentclass{article}  
\usepackage{iclr2025_conference,times}

\usepackage{amsmath,amsfonts,bm}

\def\eqref#1{equation~\ref{#1}}

\def\1{\bm{1}}

\DeclareMathAlphabet{\mathsfit}{\encodingdefault}{\sfdefault}{m}{sl}
\SetMathAlphabet{\mathsfit}{bold}{\encodingdefault}{\sfdefault}{bx}{n}

\usepackage{capt-of, eucal}
\usepackage{xspace}
\usepackage{xcolor}
\usepackage{enumitem}
\usepackage{amsmath,amsthm,amssymb,amsfonts,dsfont,pifont,bm,bbm,mathrsfs,mathtools,nicefrac,extarrows,relsize}
\usepackage{algorithm,algpseudocode,listings}
\usepackage{booktabs,multirow,adjustbox,diagbox,threeparttable,tabularray,setspace}

\definecolor{citeblue}{rgb}{0.21,0.49,0.74}
\usepackage[pagebackref=false,breaklinks,colorlinks,citecolor=citeblue,bookmarks=false]{hyperref}
\usepackage{subfigure}
\usepackage{wrapfig}
\usepackage[capitalize]{cleveref}  

\crefname{section}{Sec.}{Secs.}
\Crefname{section}{Section}{Sections}
\crefname{appendix}{App.}{Apps.}
\Crefname{appendix}{Appendix}{Appendices}
\crefname{table}{Tab.}{Tabs.}
\Crefname{table}{Table}{Tables}
\crefname{figure}{Fig.}{Figs.}
\Crefname{figure}{Figure}{Figures}
\crefname{equation}{Eq.}{Eqs.}
\Crefname{equation}{Equation}{Equations}
\crefname{theorem}{Thm.}{Thms.}
\Crefname{theorem}{Theorem}{Theorems}
\crefname{lemma}{Lem.}{Lems.}
\Crefname{lemma}{Lemma}{Lemmas}
\crefname{remark}{Rem.}{Rems.}
\Crefname{remark}{Remark}{Remarks}
\crefname{corollary}{Cor.}{Cors.}
\Crefname{corollary}{Corollary}{Corollaries}
\crefname{algorithm}{Alg.}{Algs.}
\Crefname{algorithm}{Algorithm}{Algorithms}
\definecolor{cellred}{RGB}{213, 123, 101}
\definecolor{cellgreen}{RGB}{0, 205, 0}
\definecolor{cellblue}{RGB}{54, 125, 189}
\definecolor{codegreen}{rgb}{0,0.6,0}
\definecolor{codegray}{rgb}{0.5,0.5,0.5}
\definecolor{codepurple}{rgb}{0.58,0,0.82}
\definecolor{backcolour}{rgb}{1.0,1.0,1.0}
\lstdefinestyle{mystyle}{
    backgroundcolor=\color{backcolour},
    commentstyle=\color{codegreen},
    keywordstyle=\color{magenta},
    numberstyle=\tiny\color{codegray},
    stringstyle=\color{codepurple},
    basicstyle=\ttfamily\scriptsize,
    breakatwhitespace=false,
    breaklines=true,
    captionpos=b,
    keepspaces=true,
    numbers=left,
    numbersep=5pt,
    showspaces=false,
    showstringspaces=false,
    showtabs=false,
    tabsize=2
}
\definecolor{demphcolor}{gray}{.2}

\definecolor{demphcolorinline}{gray}{.3}

\definecolor{demphcolor1}{gray}{.6}
\newcommand{\demphs}[1]{\textcolor{demphcolor1}{#1}}

\newcommand{\tocite}[1]{{\color{red} [TO CITE]}}
\newcommand{\methodname}{Framer}
\newcommand{\task}{{video frame interpolation}\xspace}
\newcommand{\method}{\mbox{\texttt{\methodname}}\xspace}

\newcommand{\StartMenu}{\raisebox{-0.18cm}{\includegraphics[scale=0.025]{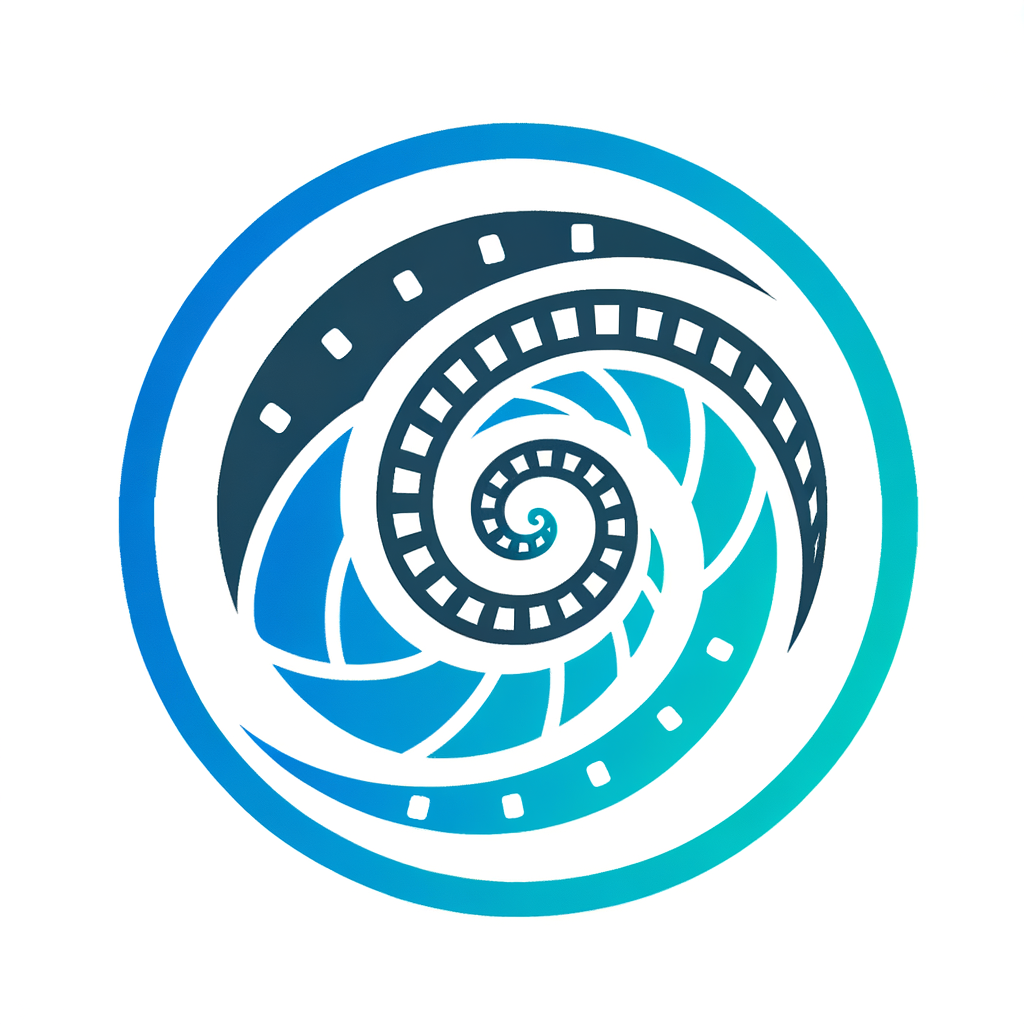}}}%
\title{\StartMenu~ \methodname: Interactive Frame Interpolation}

\author{Wen Wang$^{1,2}$,
Qiuyu Wang$^{2}$,
Kecheng Zheng$^{2}$,
Hao Ouyang$^{2}$, 
Zhekai Chen$^{1}$,
Biao Gong$^{2}$,\\
\textbf{Hao Chen$^{1}$, Yujun Shen$^{2}$, Chunhua Shen$^{1}$}\vspace{0.3cm}\\
$^1$Zhejiang University\quad\quad 
$^2$Ant Group\\
}

\iclrfinalcopy 

\begin{document}

\maketitle

\begin{figure*}[h]
\centering
\includegraphics[width=\textwidth]{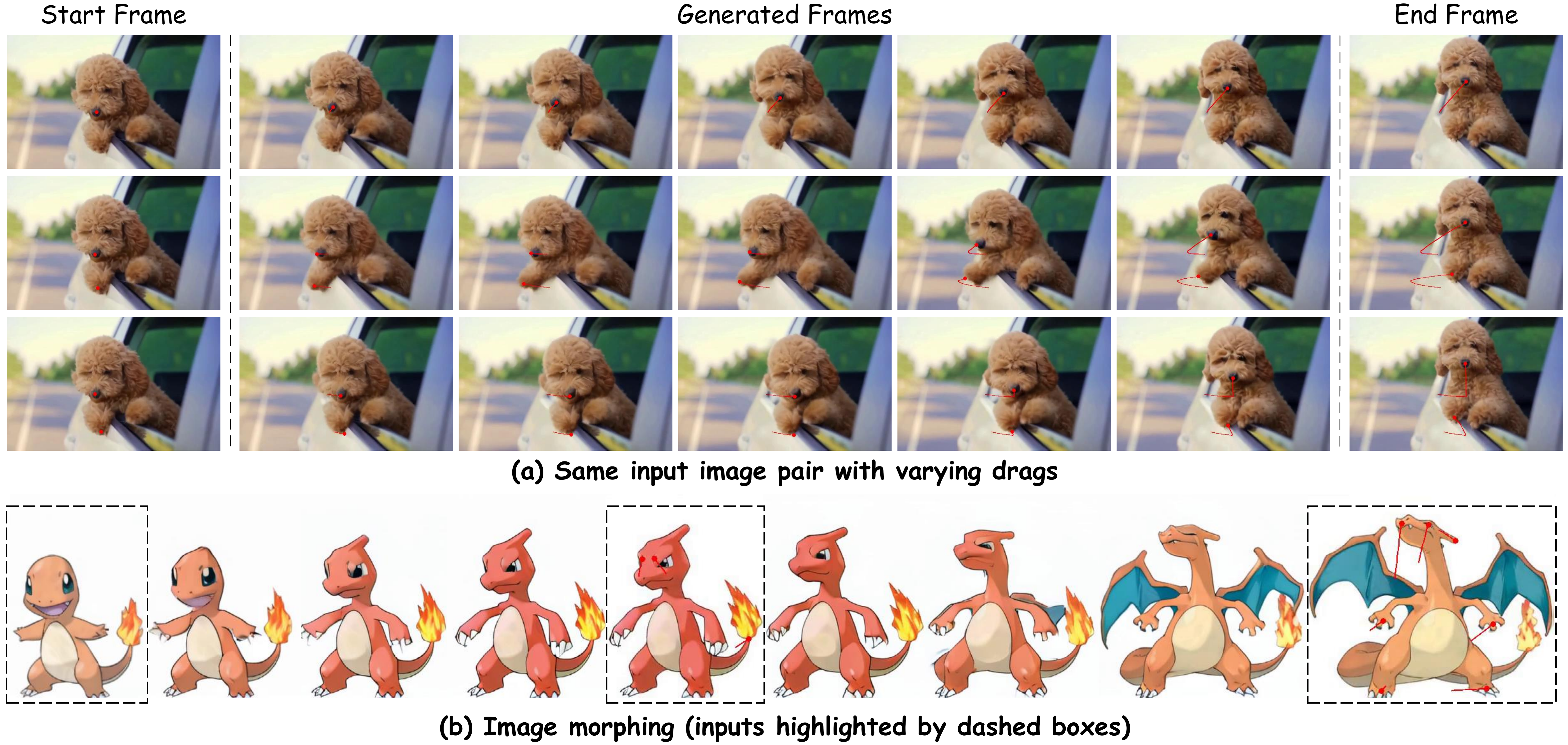}
   \caption{Showcases produced by our \method. It facilitates fine-grained customization of local motions and generates varying interpolation results given the same input start and end frame pair (first 3 rows). Moreover, \method handles challenging cases and can realize smooth image morphing (last 2 rows). The input trajectories are overlayed on the frames.}
    \label{fig:teaser}
\end{figure*}

\begin{abstract}

We propose \method for interactive frame interpolation, which targets producing smoothly transitioning frames between two images as per user creativity.
Concretely, besides taking the start and end frames as inputs, our approach supports customizing the transition process by tailoring the trajectory of some selected keypoints.
Such a design enjoys two clear benefits.
First, incorporating human interaction mitigates the issue arising from numerous possibilities of transforming one image to another, and in turn enables finer control of local motions.
Second, as the most basic form of interaction, keypoints help establish the correspondence across frames, enhancing the model to handle challenging cases (\textit{e.g.}, objects on the start and end frames are of different shapes and styles).
It is noteworthy that our system also offers an ``autopilot'' mode, where we introduce a module to estimate the keypoints and refine the trajectory automatically, to simplify the usage in practice.
Extensive experimental results demonstrate the appealing performance of \method on various applications, such as image morphing, time-lapse video generation, cartoon interpolation, \textit{etc.}
The code, the model, and the interface will be released to facilitate further research.

Project page:  \href{https://aim-uofa.github.io/Framer}{ $\tt aim$-$\tt uofa.github.io/Framer$}

\end{abstract}

\section{Introduction}

The creation of seamless and visually appealing transitions between frames~\citep{dong2023video} is a fundamental requirement in various applications, including image morphing~\citep{image_morphing}, slow-motion video generation~\citep{reda2022film}, and cartoon interpolation~\citep{xing2024tooncrafter}. Users often need to control the motion trajectories, deformation dynamics, and temporal coherence of interpolated frames to achieve specific outcomes. Therefore, incorporating interactive capabilities into frame interpolation frameworks is crucial for expanding the practical applicability.

Traditional video frame interpolation methods~\citep{superslomo2018, qvi2019, eqvi2020, softmaxsplat2020, xvfi2021, adacof2020, cdfi2021} often rely on estimating optical flow or motion to predict intermediate frames deterministically.
While significant progress has been made in this area, these approaches struggle in scenarios involving large motion or substantial changes in object appearance, due to an inaccurate flow estimation. 
What's more, when transforming one image to another, there can be numerous plausible ways objects and scenes can transition. A deterministic result may not align with user expectations or creative intent.

Orthogonal to existing methods, we propose \method, an interactive frame interpolation framework designed to produce smoothly transitioning frames between two images.
Our approach allows users to customize the transition process by tailoring the trajectories of selected keypoints, thus directly influencing the motion and deformation of objects within the scene. 
Such design offers two significant benefits.
\textbf{First}, the incorporation of keypoint-based interaction resolves the ambiguity inherent in transforming one image into another, allowing for precise control over how specific regions of the image move and change. As shown in \cref{fig:teaser}a, users can control the movements of the dog's paw and head through simple and intuitive interactions.
\textbf{Second}, keypoint trajectories establish explicit correspondences across frames, which is especially beneficial in challenging cases where objects change in shape, style, or even semantic meaning.
As shown in \cref{fig:teaser}b, the keypoint trajectories establish the correspondences between keypoints from Pokémon in varying forms and help produce a smooth ``evolution'' process of Pokémon.

Concretely, we view \task from a generative perspective and finetune a large-scale pre-trained image-to-video diffusion model~\citep{blattmann2023stable} on open-domain video datasets~\citep{nan2024openvid} to facilitate \task.
The additional last-frame conditioning is introduced during the fine-tuning process.
Afterward, a point trajectory controlling branch is introduced to take the additional point trajectory inputs, thus guiding the video interpolation process.
During inference, \method supports the ``interactive'' mode for customized \task, following user-input point trajectories.

Understanding that manual keypoint annotation may not always be desirable, we offer an ``autopilot'' mode for \method. 
Technically, we propose a novel bi-directional point-tracking method that estimates the trajectories of matched points over the entire video sequence, by analyzing both forward and backward motions between frames. 
It automates the process of obtaining keypoint trajectories, enabling \method to generate motion-natural and temporally coherent interpolation results without requiring extensive user input. 
The ``autopilot'' mode simplifies the workflow while still benefiting from the enhanced correspondence provided by the points trajectories.

We conduct extensive experiments to evaluate the performance of \method across various applications, including image morphing, time-lapse video generation, and cartoon interpolation. The results demonstrate that \method produces smooth and visually appealing transitions, outperforming existing methods, particularly in cases involving complex motions and significant appearance changes. By combining the strengths of generative models with user-guided interactions, \method improves both the quality and controllability of the interpolated frames.

\section{Related Work}

\subsection{Video Frame Interpolation}
Video frame interpolation (VFI) aims to synthesize intermediate frames from two successive video frames. Most previous methods view VFI as a low-level task, assuming a moderate motion between frames. 
These methods can roughly be categorized as flow-based methods and kernel-based methods. Specifically, the flow-based methods leverage estimated optical flow for frame synthesis~\citep{superslomo2018, qvi2019, eqvi2020, softmaxsplat2020, contextaware2018, xvfi2021, huang2020rife, Jin_2023_WACV, vimeo, bmbc, abme2021, ifrnet}. 
By contrast, the kernel-based methods rely on spatially adaptive kernels to synthesize the interpolated pixels~\citep{adacof2020, edsc2020, cdfi2021, sepconv2017, dsepconv2020, featureflow, vfiformer}. 
While the former potentially suffers from inaccurate flow estimation, the latter are often constrained by kernel size. 
To obtain the best of both worlds, some methods combine the flow- and kernel-based methods for end-to-end video frame interpolation~\citep{dain2019, memc2021, Danier_2022_CVPR,hvfi}. 
We refer readers to \citep{dong2023video} for a more comprehensive survey on these methods. 

Recently, inspired by the generative capacity of large-scale pre-trained video diffusion models, some methods attempt to tackle VFI from a generation perspective \citep{danier2024ldmvfi, feng2024explorative, jain2024video, xing2023dynamicrafter, wang2024generative}. 
For example, LDMVFI~\citep{danier2024ldmvfi} formulates VFI as a conditional generation problem and utilizes a latent diffusion model for perceptually oriented video frame interpolation. 
Similarly, VIDIM~\citep{jain2024video} leverages cascaded diffusion models to generate high-fidelity interpolated videos with nonlinear motions.
Though progress has been made, these methods still have difficulties in tackling large differences between the starting and ending frames. 
Moreover, they focus on generating a single deterministic solution for video frame interpolation, without controllability. 
Differently, we can generate multiple plausible solutions under large motion changes, and allow simple and intuitive drag interaction for user-intended interpolation results.

\subsection{Video Diffusion Models}

Large-scale pre-trained video diffusion models \citep{Sora, AlignYourLatents, PYoCo, VideoCrafter1, VideoCrafter2, wang2023modelscope, blattmann2023stable} have shown unprecedented generation results in visual quality, diversity, and realism. 
These methods leverage text or starting image controls, which are often insufficient in precision and interactiveness.  
Inspired by the success in controllable image generation \citep{ControlNet, mou2024t2i}, several works attempt to add additional controls to video diffusion models. 
Early explorations \citep{wang2024videocomposer, guo2023sparsectrl} utilize structural controls, like sketch and depth maps, for video generation. However, these control signals are often difficult to obtain during sampling, limiting their practical applications.
Differently, recent works focus on motion control and introduce trajectory control for object motion~\citep{wu2024draganything, mou2024revideo, yin2023dragnuwa} and camera pose control for camera motion~\citep{MotionCtrl, CameraCtrl, bahmani2024vd3d}. 
Both control signals can be obtained through easy and intuitive user interactions.
In this paper, we enhance the creative potential and flexibility of the video framer interpolation process, allowing users to produce plausible frame interpolation results following their control.

\section{Method}

\begin{figure*}
    \centering
    \includegraphics[width=\textwidth]{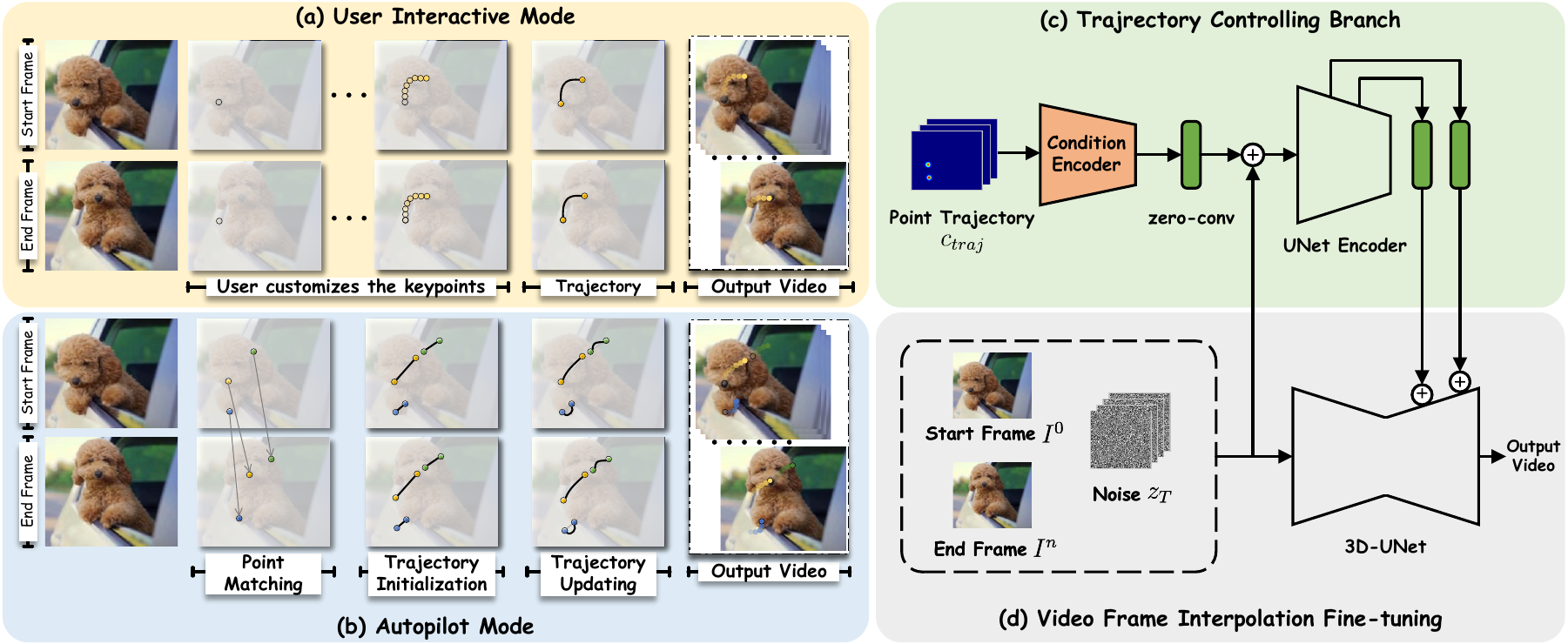}
    \caption{%
        \method supports (a) a user-interactive mode for customized point trajectories and (b) an ``autopilot'' mode for \task without trajectory inputs.
        During training, (d) we fine-tune the 3D-UNet of a pre-trained video diffusion model for \task. Afterward, (c) we introduce point trajectory control by freezing the 3D-UNet and fine-tuning the controlling branch.
    }
    \label{fig:method_main}
\end{figure*}

Given two frames, $I^0$ and $I^n$, indicating the start and end frame in a video, our goal is to generate the plausible contiguous video $I = \{I^i\}_{i=0}^{n}$ by sampling from the conditional distribution $p\left(I \mid I^0, I^n\right)$.
Here, $n$ is the number of frames in the video.
Our method, termed \method, supports a user-interactive mode for customized point trajectories and an ``autopilot'' mode for \task without trajectory inputs, as shown in \cref{fig:method_main}a and \cref{fig:method_main}b.
In the following, we will introduce how we add frame conditions to the video diffusion model to achieve video interpolation in \cref{subsec:method_arch}. 
To support user-interactive drag control, we introduce a control branch in \cref{subsec:method_corres} for point trajectory guidance, which also enhances point correspondences across frames.
In the ``autopilot'' mode, we estimate trajectories of matched points in the video with our novel bi-directional point tracking method, as illustrated in \cref{subsec:method_track}.

\subsection{Model Architecture}
\label{subsec:method_arch}

Large-scale pre-trained video diffusion models have a strong visual prior on the appearance, structure, and movement of open-world objects~\citep{Sora}. Our approach builds on the video diffusion model to exploit this prior. Considering that the Image-to-Video (I2V) diffusion model naturally supports first-frame conditioning, we choose the representative I2V diffusion model, Stable Video Diffusion (SVD)~\citep{blattmann2023stable}, as our base model, as shown in \cref{fig:method_main}d.

Based on the I2V model, we need to introduce additional end-frame conditioning to realize video interpolation. To preserve the visual prior of the pre-trained SVD as much as possible, we follow the conditioning paradigm of SVD and inject end-frame conditions in the latent space and semantic space, respectively. 
Specifically, we concatenate the VAE-encoded latent feature of the first frame, denoted as $z^0$, with the noisy latent of the first frame, as did in SVD. Additionally, we concatenate the latent feature of the last frame, $z^n$, with the noisy latent of the end frame, considering that the conditions and the corresponding noisy latents are spatially aligned. 
In addition, we extract the CLIP image embedding of the first and last frames separately and concatenate them for cross-attention feature injection.
The U-Net $\epsilon_\theta$ is trained using the denoising score matching objective:
\begin{equation}
\mathcal{L}=\mathbb{E}_{z_t, z^0, z^n, t, \epsilon \sim \mathcal{N}(0, \mathbf{I})}\left[\left\|\epsilon-\epsilon_\theta\left(z_t; t, z^0, z^n\right)\right\|^2\right].
\end{equation}

\subsection{Interactive Frame Interpolation}
\label{subsec:method_corres}

Ambiguity remains given the start and end frames, especially when the distinction between the two frames is large.
The reason is that multiple plausible interpolation results can be obtained by sampling video from the conditional distribution $P \left(I \mid I^0, I^n\right)$ for the same input pair. 
To better align with the user intention, we introduce a control branch for custmized point trajectory guidance.

Technically, we train a point trajectory-based control branch for correspondence modeling, as shown in \cref{fig:method_main}c. During training, we use the following steps to obtain the point trajectory as control signals. Firstly, we randomly initialize some sampled points around a fixed sparse grid in the first frame, and use Co-Tracker~\citep{karaev2023cotracker} to obtain the trajectories of these points in the whole video. Secondly, we remove trajectories that are not visible in more than half of the video frames. Lastly, we sample the point trajectories with larger motions with greater probability. Considering that the users usually only input a small number of point trajectories, we keep only 1 to 10 trajectories during training. 
Please refer to the \cref{subsec:app_implemtation} for more details.

After obtaining the sampled point trajectories, we follow DragNUWA~\citep{yin2023dragnuwa} and DragAnything~\citep{wu2024draganything} to transform the point coordinates into a Gaussian heatmap, denoted as $c_{traj}$, which is used as input to the control module. 
We follow the conditioning mechanism in ControlNet~\citep{ControlNet} to incorporate the trajectory control. Specifically, we copy the encoder of 3D-UNet to encode the trajectory map and add it into the decoder of U-Net after zero-convolution~\citep{ControlNet}. This training process can be represented as:
\begin{equation}
\mathcal{L}=\mathbb{E}_{z_t, z^0, z^n, t, \epsilon \sim \mathcal{N}(0, \mathbf{I})}\left[\left\|\epsilon-\epsilon^{c}_{\theta}\left(z_t; t, z^0, z^n, c_{traj}\right)\right\|^2\right].
\label{eq:conditonal_denoising}
\end{equation}
Here, $\epsilon^{c}_{\theta}$ is the combination of the denoising U-Net and the ControlNet branch.

\textbf{Discussion.}
The introduction of point trajectory control not only facilitates user interaction, but also enhances the correspondence among points from different frames.
As demonstrated in experiments, this approach enables the model to effectively tackle challenging cases, such as when the start and end frames differ significantly.

\subsection{``Autopilot'' Mode for Frame Interpolation}
\label{subsec:method_track}

In practical applications, users may not always prefer manual drag controls. For this reason, we propose an ``autopilot'' mode to enhance the ease of use of our \method.
It mainly contains a trajectory initialization and a trajectory updating process, as illustrated in \cref{fig:method_main}b.

\textbf{Trajectory Initialization.}
Given the start and end frames of the input video, we can obtain the matching points between the two frames by applying feature-matching algorithms. 
The matched points are denoted as $\{\boldsymbol{p}_i\}^{m}_{i=1}$, where $m$ is the number of matching points. $\boldsymbol{p}_i$ denotes the known anchor points on the trajectory. At initialization, it contains the matched points on the first and last frames, \textit{i.e.}, $\boldsymbol{p}_i = [p^0_i, p^n_i]$.
Although varying feature matching algorithms are feasible, we use the classical SIFT feature matching~\citep{lowe2004distinctive} here for its simplicity and effectiveness.
Subsequently, we can obtain the $i$-th trajectory $\hat{c}_{i}$ by interpolating the anchor points $\boldsymbol{p}_i$. 
The estimated trajectory for all $m$ matched points, denoted as $\hat{c}_{traj} = \{\hat{c}_{i}\}^m_{i=1} $, are used as the input condition in \cref{eq:conditonal_denoising}.

\begin{figure*}
    \centering
    \includegraphics[width=\textwidth]{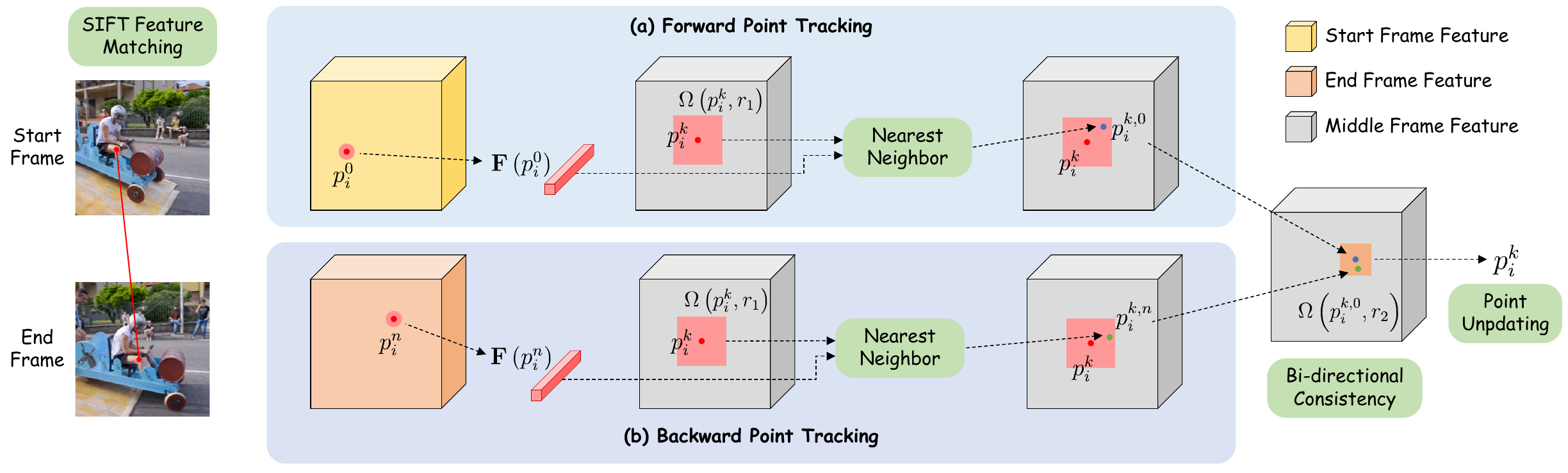}
    \caption{%
        \textbf{Point trajectory estimation.} The point trajectory is initialized by interpolating the coordinates of matched keypoints. In each de-noising step, we perform point tracking by finding the nearest neighbor of keypoints in the start and end frames, respectively. Lastly, We check the bi-directional tracking consistency before updating the point coordinate.
    }
    \label{fig:method_traj}
\end{figure*}

\textbf{Trajectory Updating.}
Although the initial trajectory provides temporally consistent point correspondence, the trajectory obtained by connecting points in the first and last frames may not be accurate. 
Inspired by DragGAN~\citep{pan2023drag} and DragDiffusion~\citep{shi2024dragdiffusion}, we perform point tracking using the intermediate feature in U-Net to update the trajectories. 
Specifically, in each denoising step, we interpolate the U-Net features to the image resolution, denoted as $\mathbf{F}$. 
Here we use the feature of the penultimate upsampled block in U-Net, since it enjoys a good trade-off between feature resolution and discriminativeness.
We use $\mathbf{F}(p)$ to represent the feature of the point $p$, which is obtained via bilinear interpolation, since the coordinates may not be integers.

In each denoising step, we apply point tracking to update the coordinates of the middle frame points.
We use nearest neighbor search in a feature patch around the point. 
The feature patch represents a set of points whose distance to point $p$ is less than $r$, and is denoted as 
$\Omega\left({p}, r\right)=\left\{(x, y)|| x-x_{p}\left|<r,\right| y-y_{p } \mid<r\right\}$. 
For a middle frame point ${p}^{k}_i$ in the $k$-th frame, we find the nearest point relative to the anchor point ${p}^{0}_i$ via:
\begin{equation}
{p}^{k,0}_i:=\underset{{q}^k_i \in \Omega\left({p}^k_i, r_1\right)}{\arg \min }\left\|\mathbf{F}\left({q}^k_i\right)-\mathbf{F}\left({p}^0_i\right)\right\|_1.
\end{equation}
Similarly, we can obtain the nearest point relative to the last anchor point ${p}^{n}_i$:
\begin{equation}
{p}^{k,n}_i:=\underset{{q}^k_i \in \Omega\left({p}^k_i, r_1\right)}{\arg \min }\left\|\mathbf{F}\left({q}^k_i\right)-\mathbf{F}\left({p}^n_i\right)\right\|_1.
\end{equation}
As shown in \cref{fig:method_traj}, to further ensure the accuracy of the coordinates of the updated points, we check the consistency of the two nearest points obtained by matching with ${p}^{0}_i$ and ${p}^{n}_i$. 
When the distance between the two is less than a threshold $r_2$, \textit{i.e.}, ${p}^{k,n}_i \in \Omega\left({p}^{k,0}_i, r_2\right)$, we update the point coordinates by setting ${p}^{k}_i = ({p}^{k,0}_i + {p}^{k, n}_i) / 2$.
Then, we add the point to the anchor points list $\boldsymbol{p}_i$ and interpolate $\boldsymbol{p}_i$ to get the updated trajectory $c_i$, which is used as the input condition to the next denoising step.

\section{Experiments}

\subsection{Implementation Details}

Our method is built on SVD and trained on the high-quality OpenVidHD-0.4M dataset~\citep{nan2024openvid}. 
During the training of U-Net, we fixed the spatial attention and residual blocks, and only fine-tuned the input convolutional and temporal attention layers. The model is trained for 100K iterations using the AdamW optimizer~\citep{loshchilov2017decoupled} with a learning rate of 1e-5. 
We obtained the point trajectories by pre-processing the video using the Co-Tracker~\citep{karaev2023cotracker}. 
When training the control module, we fixed the U-Net and optimized the control module for 10K steps using the AdamW optimizer, with a learning rate of 1e-5. 
All training is performed on 16 NVIDIA A100 GPUs, and the total batch size is 16.
During ``autopilot'' mode sampling, we keep $m=5$ best matching keypoints for trajectory guidance, and the distance thresholds for point tracking are set as $r_1=5$ and $r_2=3$.
Please refer to \cref{subsec:app_implemtation} for more details.

\subsection{Comparison}

\begin{figure*}
    \centering
    \includegraphics[width=\textwidth]{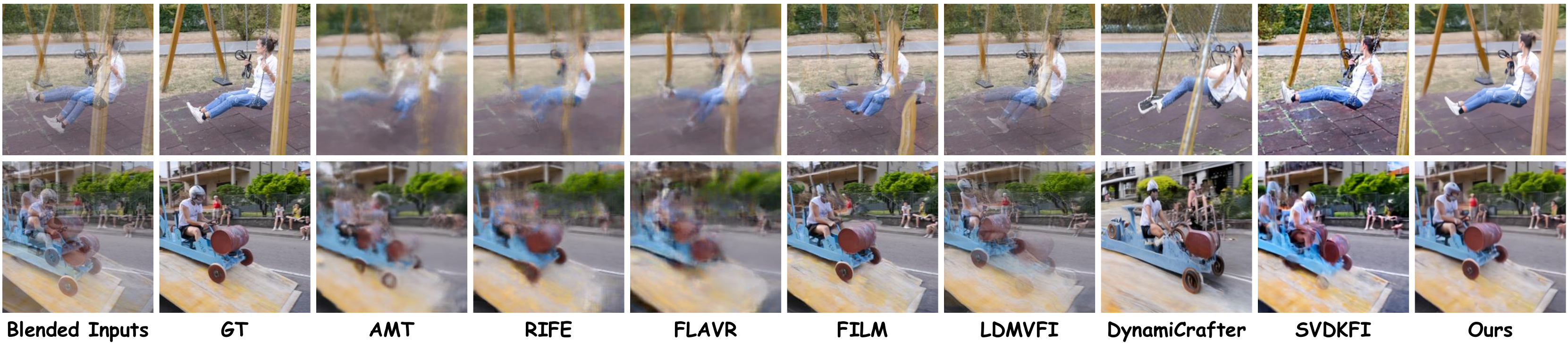}
    \caption{%
        Qualitative comparison. 
        `GT" strands for ground truth.
        For each method, we only present the middle frame of 7 interpolated frames. The full results can be seen in \cref{fig:app_comparison_1} and \cref{fig:app_comparison_2} in the Appendix.
    }
    \label{fig:results_compare}
\end{figure*}

\begin{wrapfigure}{r}{6cm}
\centering
\includegraphics[width=0.25\textwidth]{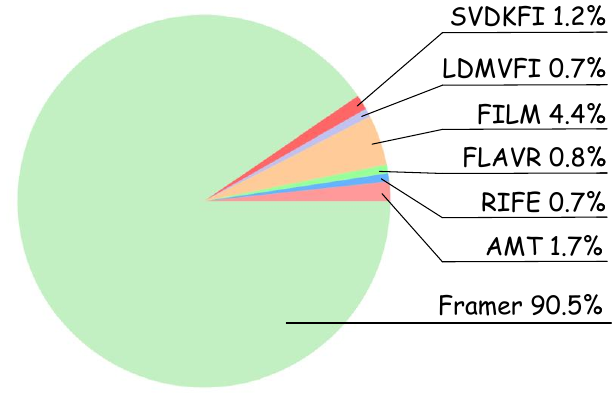}
\caption{Reults on human preference.}
\label{fig:user_study}
\end{wrapfigure}

Existing methods do not support drag-user interaction. Thus, we use the ``autopilot'' mode of \method to make fair comparisons.
We select baselines from two distinct categories. The first category includes the latest general diffusion-based video interpolation models, including LDMVFI~\citep{danier2024ldmvfi}, DynamicCrafter~\citep{xing2023dynamicrafter}, and SVDKFI~\citep{wang2024generative}. The second category encompasses traditional video interpolation methods, such as AMT~\citep{licvpr23amt}, RIFE~\citep{huang2020rife}, FLAVR \citep{flavr}, and FILM~\citep{reda2022film}. 
We conduct quantitative and qualitative analyses, as well as user studies, on two publicly available datasets: DAVIS~\citep{pont20172017} and UCF101~\citep{ucf101}.

\textbf{Qualitative Comparison.}
As shown in \cref{fig:results_compare}, our method produces significantly clearer textures and natural motion compared to existing interpolation techniques. 
It performs especially well in scenarios with substantial differences between the input frames, where traditional methods often fail to interpolate content accurately. Compared to other diffusion-based methods like LDMVFI and SVDKFI, \method demonstrates superior adaptability to challenging cases and offers better control. 

\textbf{Quantitative Comparison.}
As discussed in VIDIM~\citep{jain2024video}, reconstruction metrics like PSNR, SSIM, and LPIPS fail to capture the quality of interpolated frames accurately, since they penalize other plausible interpolation results that are not pixel-aligned with the original video. 
While generation metrics such as FID offer some improvement, they still fall short as they do not account for temporal consistency and evaluate frames in isolation. 
Despite this, we present the quantitative metrics for various settings on both datasets, where our method achieves the best FVD score among all baselines as in \cref{tab:compare_7}.
We also evaluate \method with 5 random point trajectories from ground-truth videos, estimated using Co-Tracker. As can be seen, ``\method with Co-Tracker'' achieves superior performance even in reconstruction metric.
For a more comprehensive assessment of quality, we recommend reviewing the supplementary comparison videos.

\begin{table*}[t]
\small
\renewcommand\tabcolsep{5.0pt}
\centering
\resizebox{\textwidth}{!}{
\begin{tabular}[t]{rccccc|lcccc}
&  \multicolumn{5}{c}{\textbf{DAVIS-7}}{\hskip 0.5cm} & \multicolumn{5}{c}{\textbf{UCF101-7}}{\hskip 0.25cm}  \\
\cline{2-11}
  & PSNR$\uparrow$ & SSIM$\uparrow$ & LPIPS$\downarrow$ & FID$\downarrow$ & FVD$\downarrow$ & PSNR$\uparrow$ & SSIM$\uparrow$ & LPIPS$\downarrow$ & FID$\downarrow$ & FVD$\downarrow$ \\ 
\hline
AMT \citep{licvpr23amt}                         & 21.66 & \textbf{0.7229} & 0.2860 & 39.17 & 245.25 & 26.64 & 0.9000 & 0.1878 & 37.80 & 270.98  \\
RIFE \citep{huang2020rife}                      & \textbf{22.00} & 0.7216 & 0.2663 & 39.16 & 319.79 & \textbf{27.04} & \textbf{0.9020} & 0.1575 & 27.96 & 300.40  \\
FLAVR \citep{flavr}                              & 20.94 & 0.6880 & 0.3305 & 52.23 & 296.37 & 26.50 & 0.8982 & 0.1836 & 37.79 & 279.58  \\ 
FILM \citep{reda2022film}                       & 21.67 & 0.7121 & \textbf{0.2191} & \textbf{17.20} & 162.86 & 26.74 & 0.8983 & \textbf{0.1378} & \textbf{16.22} & 239.48  \\
LDMVFI \citep{danier2024ldmvfi}                 & 21.11 & 0.6900 & 0.2535 & 21.96 & 269.72 & 26.68 & 0.8955 & 0.1446 & 17.55 & 270.33  \\
DynamicCrafter \citep{xing2023dynamicrafter}    & 15.48 & 0.4668 & 0.4628 & 35.95 & 468.78 & 17.62 & 0.7082 & 0.3361 & 61.71 & 646.91  \\
SVDKFI \citep{wang2024generative}                                 & 16.71 & 0.5274 & 0.3440 & 26.59 & 382.19 & 21.04 & 0.7991 & 0.2146 & 44.81 & 301.33  \\
\hline
\textbf{\method (Ours)}                           & 21.23 & 0.7218 & 0.2525 & 27.13 & \textbf{115.65} & 25.04 & 0.8806 & 0.1714 & 31.69 & \textbf{181.55}  \\
\demphs{\method with Co-Tracker (Ours)}                           & \demphs{22.75} & \demphs{0.7931} & \demphs{0.2199} & \demphs{27.43} & \demphs{102.31} & \demphs{27.08} & \demphs{0.9024} & \demphs{0.1714} & \demphs{32.37} & \demphs{159.87}  \\
\hline
\end{tabular}
}
\caption{
Quantitative comparison with existing video interpolation methods on reconstruction and generative metrics, evaluated on all 7 generated frames. 
}
\label{tab:compare_7}
\end{table*}

\begin{figure*}
    \centering
    \includegraphics[width=\textwidth]{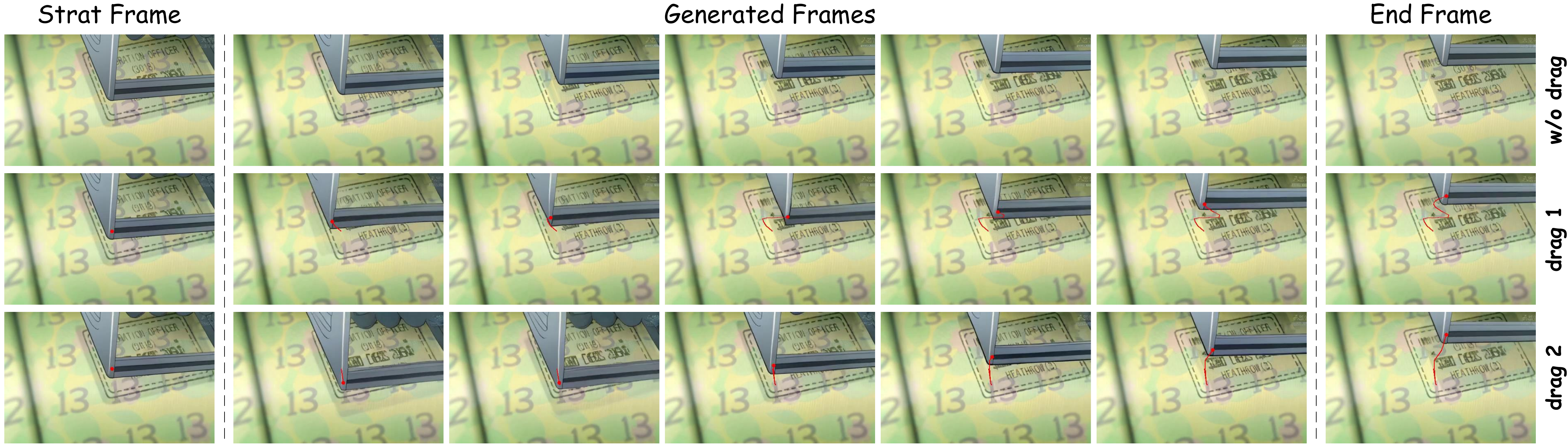}
    \caption{%
        {Results on user interaction.} The first row is generated without drag input, while the other two are generated with different drag controls. Customized trajectories ares overlaid on frames.
    }
    \label{fig:results_drag}
\end{figure*}

\begin{figure*}
    \centering
    \includegraphics[width=\textwidth]{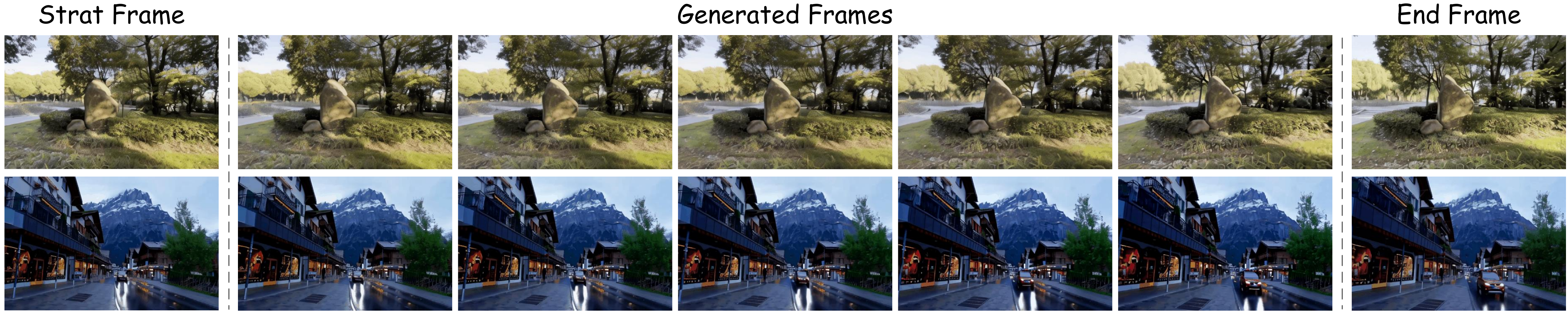}
    \caption{%
        Novel view synthesis on both static (1st row) and dynamic scenes (2nd row).
    }
    \label{fig:results_nvs}
\end{figure*}

\textbf{User Study.}
Since quantitative metrics fall short in reflecting video quality, we further assessed our method's performance through a user study. In this study, participants reviewed video sets generated from the same input frame pair by both existing methods and our \method. Participants assessed up to 100 randomly ordered video sets and selected the one they found most realistic. In total, 20 participants provided 1,000 ratings across these video sets. As illustrated in \cref{fig:user_study}, the results demonstrate a strong preference among human raters for the outputs produced by our method.

\subsection{Applications}

\textbf{Optional drag control.}
Given the same input start and end frames, multiple plausible results can satisfy the goal of video interpolation. 
With \method, users can direct the motion of the entities in input frames with simple drags for their intention, or simply obtain a default interpolation result without drags.
As shown in \cref{fig:results_drag}, the seal moves in varying directions given the same input frames.

\begin{figure*}
    \centering
    \includegraphics[width=\textwidth]{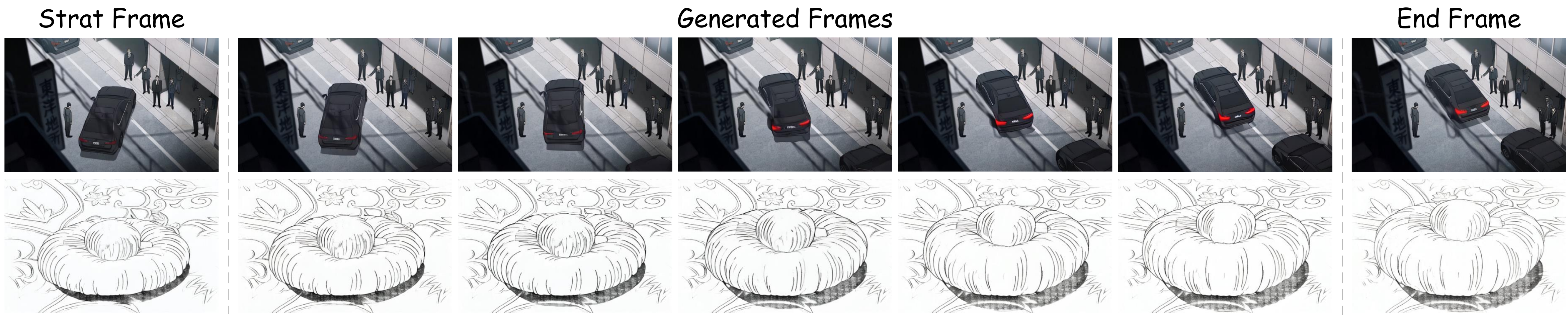}
    \caption{%
        Applications on cartoon (1st row) and sketch (2nd row) interpolation.
    }
    \label{fig:results_toon}
\end{figure*}

\begin{figure*}
    \centering
    \includegraphics[width=\textwidth]{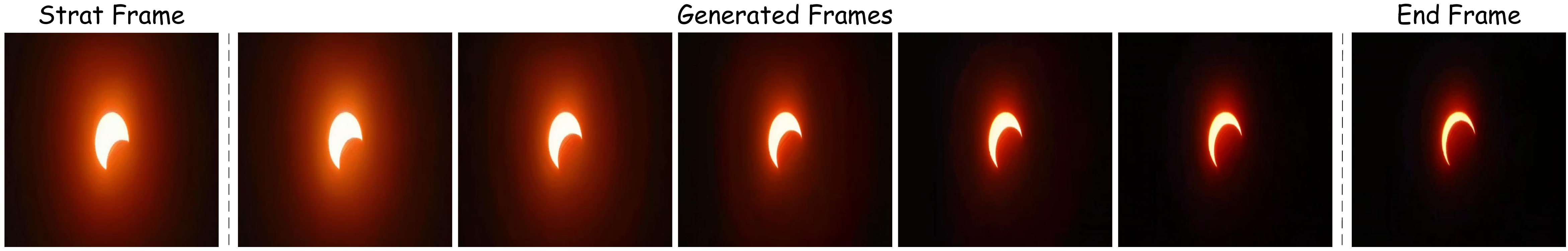}
    \caption{%
        Applications on time-lapsing video generation.
    }
    \label{fig:results_chron}
\end{figure*}

\textbf{Novel view synthesis (NVS)} is a classical 3D vision task, with a wide range of applications. 
Using images of different viewpoints as the start and end frames of the video respectively, we can realize the NVS from sparse viewpoint input by performing video interpolation. 
As shown in \cref{fig:results_nvs}, our method achieves pleasing NVS in both static scenes (first row) and dynamic scenes (second and third rows). 
Taking the second row as an example, the house gradually moves out of the scene as the camera keeps moving forward. In the meantime, the car moves in the opposite direction to the camera and gradually takes up a larger proportion in the frame.

\begin{figure*}
    \centering
    \includegraphics[width=\textwidth]{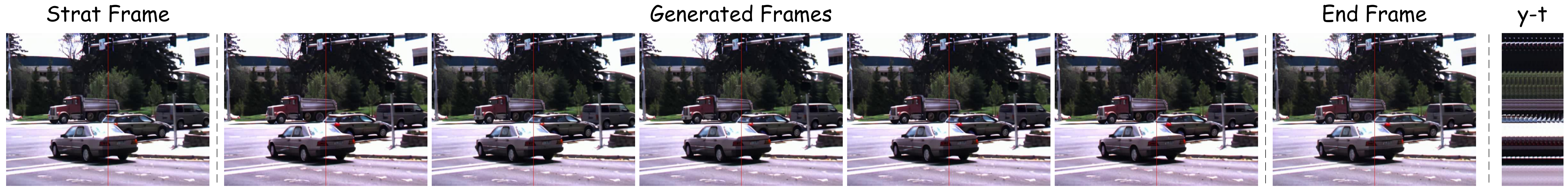}
    \caption{%
        Applications on slow-motion video generation. The y-t slice highlighted in red on video frames is visualized on the right.
    }
    \label{fig:results_slow}
\end{figure*}
\textbf{Cartoon and sketch interpolation.}
We can dramatically simplify the process of cartoon video production, by interpolating manually created cartoon images. To this end, we tested our method on cartoon data. 
Although our method is not specifically trained on cartoon videos, it produces appealing cartoon video results and supports both color images and sktech drawing frame interpolation, as shown in \cref{fig:results_toon}. 
For example, our method successfully models the motion of two objects, \textit{i.e.}, the front vehicle pulls sideways while the rear vehicle follows, as shown in the first row.
In the third row, \method produces a smooth motion of the hand lifting in sketch drawings.

\textbf{Time-lapsing video generation.}
Time-lapse photography can vividly demonstrate slow changes that are difficult to detect with the naked eye. 
Typically, it requires sufficient storage space to hold a large amount of image data and a complex post-processing procedure to organize and edit the images.
Video interpolation provides a simple and effective way to obtain time-lapse videos by interpolating frames with only a few images of key moments. 
As shown in \cref{fig:results_chron}, \method produces the smooth change of moon waxing and waning.

\textbf{Slow-motion video generation} enhances visual effects by highlighting fine details and allows closer examination of fast phenomena. Our \method inherently supports fast frame interpolation, as demonstrated in \cref{fig:results_slow}, enabling smooth slow-motion effects suitable for films and animations.

\begin{figure*}
    \centering
    \includegraphics[width=\textwidth]{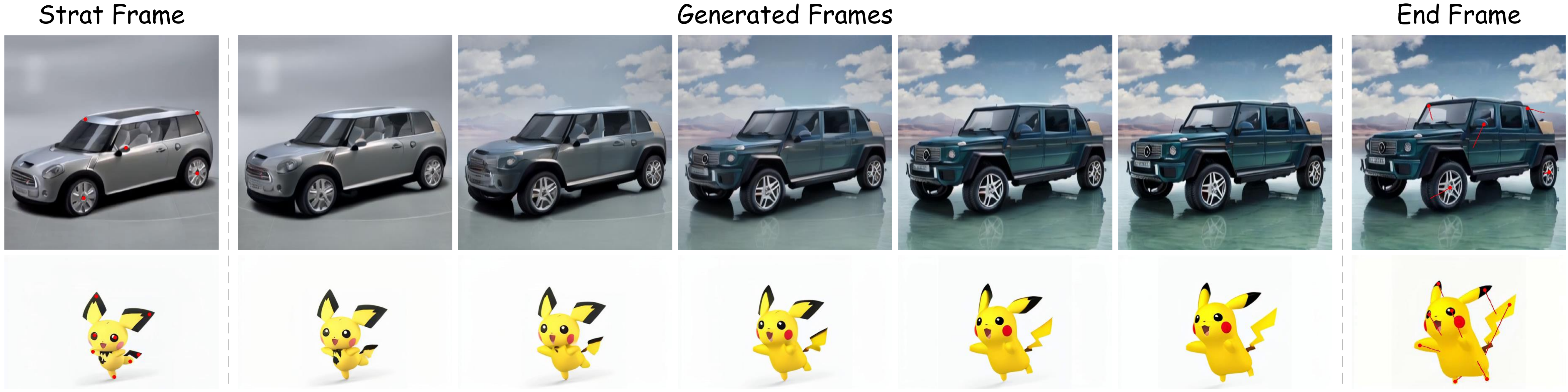}
    \caption{%
        Applications on image morphing.
        Customized trajectories ares overlaid on end frames.
    }
    \label{fig:results_morph}
\end{figure*}

\textbf{Image morphing}~\citep{image_morphing} is a popular image transformation technique with many applications in computer vision and computer graphics. Given two topologically similar images, it aims to generate a series of reasonable intermediate images. Using tue two images as the start and end frames, \method can produce natural and smooth image morphing results. 
For example, in \cref{fig:teaser}, we show the ``evolution'' process of Pokemon. More cases can be found in \cref{fig:app_morph}.

\subsection{Ablations Studies}

\begin{figure*}
    \centering
    \includegraphics[width=0.9\textwidth]{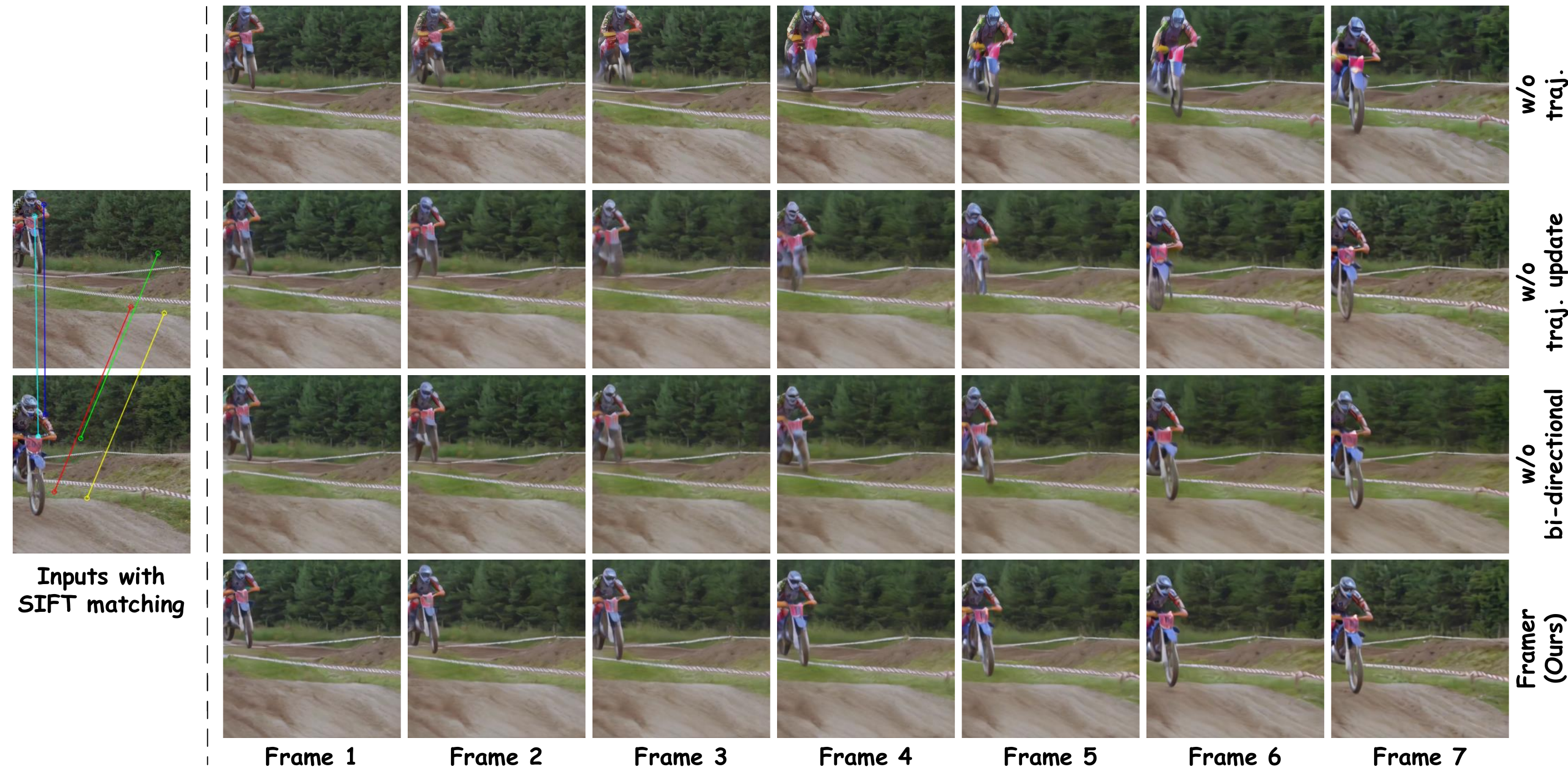}
    \caption{%
        Ablations on each component. 
        ``w/o trajectory" denotes inference without guidance from point trajectory, ``w/o traj. update" indicates inference without trajectory updates, and ``w/o bi" suggests trajectory updating without bi-directional consistency verification.
    }
    \label{fig:results_ablations}
\end{figure*}

We conducted ablation studies on the individual components of \method to validate their effectiveness. The results are illustrated in \cref{fig:results_ablations}. 
Our observations are as follows. 
First, when the trajectory guidance is removed (denoted as “w/o traj.”), the foreground motorcycle exhibits significant distortion, as shown in the 1st row of \cref{fig:results_ablations}. 
Conversely, with the inclusion of trajectory guidance, the temporal consistency of the video is notably enhanced, as depicted in the 2nd row. 
We believe this is due to the enhancement of point correspondence modeling across frames.
Second, removing trajectory updates (denoted as “w/o traj. update”) or updating the trajectory without bi-directional consistency checks (denoted as “w/o bi-directional”) results in blurring in the wheel regions of the output video. 
We suspect the blurring is caused by the guidance of unnatural motion from inaccurate trajectories, which conflicts with the generation prior in the pre-trained diffusion model, leading to local blurring.
In contrast, our method produces video frame interpolation results with natural motion and smooth temporal coherence. 
The quantitative results in \cref{tab:ablations_7,tab:ablations_1} in \cref{subsec:app_abltions} further support these findings, showing a similar trend to the qualitative ablation experiments.

\section{Conclusion and Future Work}

In this paper, we introduce \method, an interactive frame interpolation pipeline designed to produce smoothly transitioning frames between two images, guided by user-defined point trajectories.
By harnessing user input point controls from the start and end frames, we effectively guide the video interpolation process. 
Moreover, our method offers an ``autopilot'' mode that introduces a module to automatically estimate keypoints and refine trajectories without manual input. 
Through extensive experiments and user studies, we demonstrate the superiority of our method in achieving promising results in terms of both the quality and controllability of the interpolated frames. 
However, challenges remain, particularly in transitioning between different clips. 
A potential solution involves splitting the clips into several keyframes and then interpolating these keyframes sequentially.
Future work will focus on addressing these challenges.

{
\bibliographystyle{iclr2025_conference}
\bibliography{ref}

\begin{thebibliography}{60}
\providecommand{\natexlab}[1]{#1}
\providecommand{\url}[1]{\texttt{#1}}
\expandafter\ifx\csname urlstyle\endcsname\relax
  \providecommand{\doi}[1]{doi: #1}\else
  \providecommand{\doi}{doi: \begingroup \urlstyle{rm}\Url}\fi

\bibitem[Aloraibi(2023)]{image_morphing}
Alyaa Aloraibi.
\newblock Image morphing techniques: A review.
\newblock \emph{Technium: Romanian Journal of Applied Sciences and Technology}, 2023.

\bibitem[Bahmani et~al.(2024)Bahmani, Skorokhodov, Siarohin, Menapace, Qian, Vasilkovsky, Lee, Wang, Zou, Tagliasacchi, Lindell, and Tulyakov]{bahmani2024vd3d}
Sherwin Bahmani, Ivan Skorokhodov, Aliaksandr Siarohin, Willi Menapace, Guocheng Qian, Michael Vasilkovsky, Hsin{-}Ying Lee, Chaoyang Wang, Jiaxu Zou, Andrea Tagliasacchi, David~B. Lindell, and Sergey Tulyakov.
\newblock {VD3D:} taming large video diffusion transformers for 3d camera control.
\newblock \emph{arXiv: Computing Research Repo.}, abs/2407.12781, 2024.

\bibitem[Bao et~al.(2019)Bao, Lai, Ma, Zhang, Gao, and Yang]{dain2019}
Wenbo Bao, Wei{-}Sheng Lai, Chao Ma, Xiaoyun Zhang, Zhiyong Gao, and Ming{-}Hsuan Yang.
\newblock Depth-aware video frame interpolation.
\newblock In \emph{IEEE Conf. Comput. Vis. Pattern Recog.}, 2019.

\bibitem[Bao et~al.(2021)Bao, Lai, Zhang, Gao, and Yang]{memc2021}
Wenbo Bao, Wei{-}Sheng Lai, Xiaoyun Zhang, Zhiyong Gao, and Ming{-}Hsuan Yang.
\newblock Memc-net: Motion estimation and motion compensation driven neural network for video interpolation and enhancement.
\newblock \emph{IEEE Trans. Pattern Anal. Mach. Intell.}, 2021.

\bibitem[Blattmann et~al.(2023{\natexlab{a}})Blattmann, Dockhorn, Kulal, Mendelevitch, Kilian, Lorenz, Levi, English, Voleti, Letts, Jampani, and Rombach]{blattmann2023stable}
Andreas Blattmann, Tim Dockhorn, Sumith Kulal, Daniel Mendelevitch, Maciej Kilian, Dominik Lorenz, Yam Levi, Zion English, Vikram Voleti, Adam Letts, Varun Jampani, and Robin Rombach.
\newblock Stable video diffusion: Scaling latent video diffusion models to large datasets.
\newblock \emph{arXiv: Computing Research Repo.}, abs/2311.15127, 2023{\natexlab{a}}.

\bibitem[Blattmann et~al.(2023{\natexlab{b}})Blattmann, Rombach, Ling, Dockhorn, Kim, Fidler, and Kreis]{AlignYourLatents}
Andreas Blattmann, Robin Rombach, Huan Ling, Tim Dockhorn, Seung~Wook Kim, Sanja Fidler, and Karsten Kreis.
\newblock Align your latents: High-resolution video synthesis with latent diffusion models.
\newblock In \emph{IEEE Conf. Comput. Vis. Pattern Recog.}, 2023{\natexlab{b}}.

\bibitem[Brooks et~al.(2024)Brooks, Peebles, Holmes, DePue, Guo, Jing, Schnurr, Taylor, Luhman, Luhman, Ng, Wang, and Ramesh]{Sora}
Tim Brooks, Bill Peebles, Connor Holmes, Will DePue, Yufei Guo, Li~Jing, David Schnurr, Joe Taylor, Troy Luhman, Eric Luhman, Clarence Ng, Ricky Wang, and Aditya Ramesh.
\newblock Video generation models as world simulators.
\newblock \emph{OpenAI technical reports}, 2024.

\bibitem[Chen et~al.(2023)Chen, Xia, He, Zhang, Cun, Yang, Xing, Liu, Chen, Wang, Weng, and Shan]{VideoCrafter1}
Haoxin Chen, Menghan Xia, Yingqing He, Yong Zhang, Xiaodong Cun, Shaoshu Yang, Jinbo Xing, Yaofang Liu, Qifeng Chen, Xintao Wang, Chao Weng, and Ying Shan.
\newblock Videocrafter1: Open diffusion models for high-quality video generation.
\newblock \emph{arXiv: Computing Research Repo.}, abs/2310.19512, 2023.

\bibitem[Chen et~al.(2024)Chen, Zhang, Cun, Xia, Wang, Weng, and Shan]{VideoCrafter2}
Haoxin Chen, Yong Zhang, Xiaodong Cun, Menghan Xia, Xintao Wang, Chao Weng, and Ying Shan.
\newblock Videocrafter2: Overcoming data limitations for high-quality video diffusion models.
\newblock \emph{arXiv: Computing Research Repo.}, abs/2401.09047, 2024.

\bibitem[Cheng \& Chen(2020)Cheng and Chen]{dsepconv2020}
Xianhang Cheng and Zhenzhong Chen.
\newblock Video frame interpolation via deformable separable convolution.
\newblock In \emph{Assoc. Adv. Artif. Intell.}, 2020.

\bibitem[Cheng \& Chen(2022)Cheng and Chen]{edsc2020}
Xianhang Cheng and Zhenzhong Chen.
\newblock Multiple video frame interpolation via enhanced deformable separable convolution.
\newblock \emph{IEEE Trans. Pattern Anal. Mach. Intell.}, 2022.

\bibitem[Danier et~al.(2022)Danier, Zhang, and Bull]{Danier_2022_CVPR}
Duolikun Danier, Fan Zhang, and David Bull.
\newblock St-mfnet: {A} spatio-temporal multi-flow network for frame interpolation.
\newblock In \emph{IEEE Conf. Comput. Vis. Pattern Recog.}, 2022.

\bibitem[Danier et~al.(2024)Danier, Zhang, and Bull]{danier2024ldmvfi}
Duolikun Danier, Fan Zhang, and David Bull.
\newblock {LDMVFI:} video frame interpolation with latent diffusion models.
\newblock In \emph{Assoc. Adv. Artif. Intell.}, 2024.

\bibitem[Ding et~al.(2021)Ding, Liang, Zhu, and Zharkov]{cdfi2021}
Tianyu Ding, Luming Liang, Zhihui Zhu, and Ilya Zharkov.
\newblock {CDFI:} compression-driven network design for frame interpolation.
\newblock In \emph{IEEE Conf. Comput. Vis. Pattern Recog.}, 2021.

\bibitem[Dong et~al.(2023)Dong, Ota, and Dong]{dong2023video}
Jiong Dong, Kaoru Ota, and Mianxiong Dong.
\newblock Video frame interpolation: {A} comprehensive survey.
\newblock \emph{{ACM} Trans. Multim. Comput. Commun. Appl.}, 2023.

\bibitem[Feng et~al.(2024)Feng, Ding, Xia, Niklaus, Abrevaya, Black, and Zhang]{feng2024explorative}
Haiwen Feng, Zheng Ding, Zhihao Xia, Simon Niklaus, Victoria~Fern{\'{a}}ndez Abrevaya, Michael~J. Black, and Xuaner Zhang.
\newblock Explorative inbetweening of time and space.
\newblock \emph{arXiv: Computing Research Repo.}, abs/2403.14611, 2024.

\bibitem[Ge et~al.(2023)Ge, Nah, Liu, Poon, Tao, Catanzaro, Jacobs, Huang, Liu, and Balaji]{PYoCo}
Songwei Ge, Seungjun Nah, Guilin Liu, Tyler Poon, Andrew Tao, Bryan Catanzaro, David Jacobs, Jia{-}Bin Huang, Ming{-}Yu Liu, and Yogesh Balaji.
\newblock Preserve your own correlation: {A} noise prior for video diffusion models.
\newblock In \emph{Int. Conf. Comput. Vis.}, 2023.

\bibitem[Gui et~al.(2020)Gui, Wang, Chen, and Tao]{featureflow}
Shurui Gui, Chaoyue Wang, Qihua Chen, and Dacheng Tao.
\newblock Featureflow: Robust video interpolation via structure-to-texture generation.
\newblock In \emph{IEEE Conf. Comput. Vis. Pattern Recog.}, 2020.

\bibitem[Guo et~al.(2023)Guo, Yang, Rao, Agrawala, Lin, and Dai]{guo2023sparsectrl}
Yuwei Guo, Ceyuan Yang, Anyi Rao, Maneesh Agrawala, Dahua Lin, and Bo~Dai.
\newblock Sparsectrl: Adding sparse controls to text-to-video diffusion models.
\newblock \emph{arXiv: Computing Research Repo.}, abs/2311.16933, 2023.

\bibitem[He et~al.(2024)He, Xu, Guo, Wetzstein, Dai, Li, and Yang]{CameraCtrl}
Hao He, Yinghao Xu, Yuwei Guo, Gordon Wetzstein, Bo~Dai, Hongsheng Li, and Ceyuan Yang.
\newblock Cameractrl: Enabling camera control for text-to-video generation.
\newblock \emph{arXiv: Computing Research Repo.}, abs/2404.02101, 2024.

\bibitem[Huang et~al.(2020)Huang, Zhang, Heng, Shi, and Zhou]{huang2020rife}
Zhewei Huang, Tianyuan Zhang, Wen Heng, Boxin Shi, and Shuchang Zhou.
\newblock {RIFE:} real-time intermediate flow estimation for video frame interpolation.
\newblock \emph{arXiv: Computing Research Repo.}, abs/2011.06294, 2020.

\bibitem[Jain et~al.(2024)Jain, Watson, Tabellion, Holynski, Poole, and Kontkanen]{jain2024video}
Siddhant Jain, Daniel Watson, Eric Tabellion, Aleksander Holynski, Ben Poole, and Janne Kontkanen.
\newblock Video interpolation with diffusion models.
\newblock \emph{arXiv: Computing Research Repo.}, abs/2404.01203, 2024.

\bibitem[Jiang et~al.(2018)Jiang, Sun, Jampani, Yang, Learned{-}Miller, and Kautz]{superslomo2018}
Huaizu Jiang, Deqing Sun, Varun Jampani, Ming{-}Hsuan Yang, Erik~G. Learned{-}Miller, and Jan Kautz.
\newblock Super slomo: High quality estimation of multiple intermediate frames for video interpolation.
\newblock In \emph{IEEE Conf. Comput. Vis. Pattern Recog.}, 2018.

\bibitem[Jin et~al.(2023)Jin, Wu, Shen, Chen, Chen, Koo, and Hahm]{Jin_2023_WACV}
Xin Jin, Longhai Wu, Guotao Shen, Youxin Chen, Jie Chen, Jayoon Koo, and Cheul{-}Hee Hahm.
\newblock Enhanced bi-directional motion estimation for video frame interpolation.
\newblock In \emph{IEEE Winter Conf. Appl. Comput. Vis.}, 2023.

\bibitem[Kalluri et~al.(2023)Kalluri, Pathak, Chandraker, and Tran]{flavr}
Tarun Kalluri, Deepak Pathak, Manmohan Chandraker, and Du~Tran.
\newblock {FLAVR:} flow-agnostic video representations for fast frame interpolation.
\newblock In \emph{IEEE Winter Conf. Appl. Comput. Vis.}, 2023.

\bibitem[Karaev et~al.(2023)Karaev, Rocco, Graham, Neverova, Vedaldi, and Rupprecht]{karaev2023cotracker}
Nikita Karaev, Ignacio Rocco, Benjamin Graham, Natalia Neverova, Andrea Vedaldi, and Christian Rupprecht.
\newblock Cotracker: It is better to track together.
\newblock \emph{arXiv: Computing Research Repo.}, abs/2307.07635, 2023.

\bibitem[Kong et~al.(2022)Kong, Jiang, Luo, Chu, Huang, Tai, Wang, and Yang]{ifrnet}
Lingtong Kong, Boyuan Jiang, Donghao Luo, Wenqing Chu, Xiaoming Huang, Ying Tai, Chengjie Wang, and Jie Yang.
\newblock Ifrnet: Intermediate feature refine network for efficient frame interpolation.
\newblock In \emph{IEEE Conf. Comput. Vis. Pattern Recog.}, 2022.

\bibitem[Lee et~al.(2020)Lee, Kim, Chung, Pak, Ban, and Lee]{adacof2020}
Hyeongmin Lee, Taeoh Kim, Tae{-}Young Chung, Daehyun Pak, Yuseok Ban, and Sangyoun Lee.
\newblock Adacof: Adaptive collaboration of flows for video frame interpolation.
\newblock In \emph{IEEE Conf. Comput. Vis. Pattern Recog.}, 2020.

\bibitem[Li et~al.(2022)Li, Wu, Sun, Tao, Tang, and Tai]{hvfi}
Changlin Li, Guangyang Wu, Yanan Sun, Xin Tao, Chi{-}Keung Tang, and Yu{-}Wing Tai.
\newblock {H-VFI:} hierarchical frame interpolation for videos with large motions.
\newblock \emph{arXiv: Computing Research Repo.}, abs/2211.11309, 2022.

\bibitem[Li et~al.(2023)Li, Zhu, Han, Hou, Guo, and Cheng]{licvpr23amt}
Zhen Li, Zuo{-}Liang Zhu, Linghao Han, Qibin Hou, Chun{-}Le Guo, and Ming{-}Ming Cheng.
\newblock {AMT:} all-pairs multi-field transforms for efficient frame interpolation.
\newblock In \emph{IEEE Conf. Comput. Vis. Pattern Recog.}, 2023.

\bibitem[Liu et~al.(2020)Liu, Xie, Li, Sun, Qiao, and Dong]{eqvi2020}
Yihao Liu, Liangbin Xie, Siyao Li, Wenxiu Sun, Yu~Qiao, and Chao Dong.
\newblock Enhanced quadratic video interpolation.
\newblock In \emph{Eur. Conf. Comput. Vis. Worksh.}, 2020.

\bibitem[Loshchilov \& Hutter(2019)Loshchilov and Hutter]{loshchilov2017decoupled}
Ilya Loshchilov and Frank Hutter.
\newblock Decoupled weight decay regularization.
\newblock In \emph{Int. Conf. Learn. Represent.}, 2019.

\bibitem[Lowe(2004)]{lowe2004distinctive}
David~G. Lowe.
\newblock Distinctive image features from scale-invariant keypoints.
\newblock \emph{Int. J. Comput. Vis.}, 2004.

\bibitem[Lu et~al.(2022)Lu, Wu, Lin, Lu, and Jia]{vfiformer}
Liying Lu, Ruizheng Wu, Huaijia Lin, Jiangbo Lu, and Jiaya Jia.
\newblock Video frame interpolation with transformer.
\newblock In \emph{IEEE Conf. Comput. Vis. Pattern Recog.}, 2022.

\bibitem[Mou et~al.(2024{\natexlab{a}})Mou, Cao, Wang, Zhang, Shan, and Zhang]{mou2024revideo}
Chong Mou, Mingdeng Cao, Xintao Wang, Zhaoyang Zhang, Ying Shan, and Jian Zhang.
\newblock Revideo: Remake a video with motion and content control.
\newblock \emph{arXiv: Computing Research Repo.}, abs/2405.13865, 2024{\natexlab{a}}.

\bibitem[Mou et~al.(2024{\natexlab{b}})Mou, Wang, Xie, Wu, Zhang, Qi, and Shan]{mou2024t2i}
Chong Mou, Xintao Wang, Liangbin Xie, Yanze Wu, Jian Zhang, Zhongang Qi, and Ying Shan.
\newblock T2i-adapter: Learning adapters to dig out more controllable ability for text-to-image diffusion models.
\newblock In \emph{Assoc. Adv. Artif. Intell.}, 2024{\natexlab{b}}.

\bibitem[Nan et~al.(2024)Nan, Xie, Zhou, Fan, Yang, Chen, Li, Yang, and Tai]{nan2024openvid}
Kepan Nan, Rui Xie, Penghao Zhou, Tiehan Fan, Zhenheng Yang, Zhijie Chen, Xiang Li, Jian Yang, and Ying Tai.
\newblock Openvid-1m: {A} large-scale high-quality dataset for text-to-video generation.
\newblock \emph{arXiv: Computing Research Repo.}, abs/2407.02371, 2024.

\bibitem[Niklaus \& Liu(2018)Niklaus and Liu]{contextaware2018}
Simon Niklaus and Feng Liu.
\newblock Context-aware synthesis for video frame interpolation.
\newblock In \emph{IEEE Conf. Comput. Vis. Pattern Recog.}, 2018.

\bibitem[Niklaus \& Liu(2020)Niklaus and Liu]{softmaxsplat2020}
Simon Niklaus and Feng Liu.
\newblock Softmax splatting for video frame interpolation.
\newblock In \emph{IEEE Conf. Comput. Vis. Pattern Recog.}, 2020.

\bibitem[Niklaus et~al.(2017)Niklaus, Mai, and Liu]{sepconv2017}
Simon Niklaus, Long Mai, and Feng Liu.
\newblock Video frame interpolation via adaptive separable convolution.
\newblock In \emph{Int. Conf. Comput. Vis.}, 2017.

\bibitem[Pan et~al.(2023)Pan, Tewari, Leimk{\"{u}}hler, Liu, Meka, and Theobalt]{pan2023drag}
Xingang Pan, Ayush Tewari, Thomas Leimk{\"{u}}hler, Lingjie Liu, Abhimitra Meka, and Christian Theobalt.
\newblock Drag your {GAN:} interactive point-based manipulation on the generative image manifold.
\newblock In Erik Brunvand, Alla Sheffer, and Michael Wimmer (eds.), \emph{SIGGRAPH}, 2023.

\bibitem[Park et~al.(2020)Park, Ko, Lee, and Kim]{bmbc}
Junheum Park, Keunsoo Ko, Chul Lee, and Chang{-}Su Kim.
\newblock {BMBC:} bilateral motion estimation with bilateral cost volume for video interpolation.
\newblock In \emph{Eur. Conf. Comput. Vis.}, 2020.

\bibitem[Park et~al.(2021)Park, Lee, and Kim]{abme2021}
Junheum Park, Chul Lee, and Chang{-}Su Kim.
\newblock Asymmetric bilateral motion estimation for video frame interpolation.
\newblock In \emph{Int. Conf. Comput. Vis.}, 2021.

\bibitem[Pont{-}Tuset et~al.(2017)Pont{-}Tuset, Perazzi, Caelles, Arbel{\'{a}}ez, Sorkine{-}Hornung, and Gool]{pont20172017}
Jordi Pont{-}Tuset, Federico Perazzi, Sergi Caelles, Pablo Arbel{\'{a}}ez, Alexander Sorkine{-}Hornung, and Luc~Van Gool.
\newblock The 2017 {DAVIS} challenge on video object segmentation.
\newblock \emph{arXiv: Computing Research Repo.}, abs/1704.00675, 2017.

\bibitem[Reda et~al.(2022)Reda, Kontkanen, Tabellion, Sun, Pantofaru, and Curless]{reda2022film}
Fitsum~A. Reda, Janne Kontkanen, Eric Tabellion, Deqing Sun, Caroline Pantofaru, and Brian Curless.
\newblock {FILM:} frame interpolation for large motion.
\newblock In \emph{Eur. Conf. Comput. Vis.}, 2022.

\bibitem[Shi et~al.(2023)Shi, Xue, Pan, Zhang, Tan, and Bai]{shi2024dragdiffusion}
Yujun Shi, Chuhui Xue, Jiachun Pan, Wenqing Zhang, Vincent Y.~F. Tan, and Song Bai.
\newblock Dragdiffusion: Harnessing diffusion models for interactive point-based image editing.
\newblock \emph{arXiv: Computing Research Repo.}, abs/2306.14435, 2023.

\bibitem[Sim et~al.(2021)Sim, Oh, and Kim]{xvfi2021}
Hyeonjun Sim, Jihyong Oh, and Munchurl Kim.
\newblock {XVFI:} extreme video frame interpolation.
\newblock In \emph{Int. Conf. Comput. Vis.}, 2021.

\bibitem[Soomro et~al.(2012)Soomro, Zamir, and Shah]{ucf101}
Khurram Soomro, Amir~Roshan Zamir, and Mubarak Shah.
\newblock {UCF101:} {A} dataset of 101 human actions classes from videos in the wild.
\newblock \emph{arXiv: Computing Research Repo.}, abs/1212.0402, 2012.

\bibitem[Wang et~al.(2023{\natexlab{a}})Wang, Yuan, Chen, Zhang, Wang, and Zhang]{wang2023modelscope}
Jiuniu Wang, Hangjie Yuan, Dayou Chen, Yingya Zhang, Xiang Wang, and Shiwei Zhang.
\newblock Modelscope text-to-video technical report.
\newblock \emph{arXiv: Computing Research Repo.}, abs/2308.06571, 2023{\natexlab{a}}.

\bibitem[Wang et~al.(2023{\natexlab{b}})Wang, Yuan, Zhang, Chen, Wang, Zhang, Shen, Zhao, and Zhou]{wang2024videocomposer}
Xiang Wang, Hangjie Yuan, Shiwei Zhang, Dayou Chen, Jiuniu Wang, Yingya Zhang, Yujun Shen, Deli Zhao, and Jingren Zhou.
\newblock Videocomposer: Compositional video synthesis with motion controllability.
\newblock In \emph{Adv. Neural Inform. Process. Syst.}, 2023{\natexlab{b}}.

\bibitem[Wang et~al.(2024{\natexlab{a}})Wang, Zhou, Curless, Kemelmacher-Shlizerman, Holynski, and Seitz]{wang2024generative}
Xiaojuan Wang, Boyang Zhou, Brian Curless, Ira Kemelmacher-Shlizerman, Aleksander Holynski, and Steven~M Seitz.
\newblock Generative inbetweening: Adapting image-to-video models for keyframe interpolation.
\newblock \emph{arXiv: Computing Research Repo.}, abs/2408.15239, 2024{\natexlab{a}}.

\bibitem[Wang et~al.(2024{\natexlab{b}})Wang, Yuan, Wang, Li, Chen, Xia, Luo, and Shan]{MotionCtrl}
Zhouxia Wang, Ziyang Yuan, Xintao Wang, Yaowei Li, Tianshui Chen, Menghan Xia, Ping Luo, and Ying Shan.
\newblock Motionctrl: {A} unified and flexible motion controller for video generation.
\newblock In \emph{SIGGRAPH}, 2024{\natexlab{b}}.

\bibitem[Wu et~al.(2024)Wu, Li, Gu, Zhao, He, Zhang, Shou, Li, Gao, and Zhang]{wu2024draganything}
Weijia Wu, Zhuang Li, Yuchao Gu, Rui Zhao, Yefei He, David~Junhao Zhang, Mike~Zheng Shou, Yan Li, Tingting Gao, and Di~Zhang.
\newblock Draganything: Motion control for anything using entity representation.
\newblock \emph{arXiv: Computing Research Repo.}, abs/2403.07420, 2024.

\bibitem[Xing et~al.(2023)Xing, Xia, Zhang, Chen, Wang, Wong, and Shan]{xing2023dynamicrafter}
Jinbo Xing, Menghan Xia, Yong Zhang, Haoxin Chen, Xintao Wang, Tien{-}Tsin Wong, and Ying Shan.
\newblock Dynamicrafter: Animating open-domain images with video diffusion priors.
\newblock \emph{arXiv: Computing Research Repo.}, abs/2310.12190, 2023.

\bibitem[Xing et~al.(2024)Xing, Liu, Xia, Zhang, Wang, Shan, and Wong]{xing2024tooncrafter}
Jinbo Xing, Hanyuan Liu, Menghan Xia, Yong Zhang, Xintao Wang, Ying Shan, and Tien{-}Tsin Wong.
\newblock Tooncrafter: Generative cartoon interpolation.
\newblock \emph{arXiv: Computing Research Repo.}, abs/2405.17933, 2024.

\bibitem[Xu et~al.(2019)Xu, Si{-}Yao, Sun, Yin, and Yang]{qvi2019}
Xiangyu Xu, Li~Si{-}Yao, Wenxiu Sun, Qian Yin, and Ming{-}Hsuan Yang.
\newblock Quadratic video interpolation.
\newblock In \emph{Adv. Neural Inform. Process. Syst.}, 2019.

\bibitem[Xue et~al.(2019)Xue, Chen, Wu, Wei, and Freeman]{vimeo}
Tianfan Xue, Baian Chen, Jiajun Wu, Donglai Wei, and William~T. Freeman.
\newblock Video enhancement with task-oriented flow.
\newblock \emph{Int. J. Comput. Vis.}, 2019.

\bibitem[Xue et~al.(2023)Xue, Song, Guo, Liu, Zong, Liu, and Luo]{xue2024raphael}
Zeyue Xue, Guanglu Song, Qiushan Guo, Boxiao Liu, Zhuofan Zong, Yu~Liu, and Ping Luo.
\newblock {RAPHAEL:} text-to-image generation via large mixture of diffusion paths.
\newblock In \emph{Adv. Neural Inform. Process. Syst.}, 2023.

\bibitem[Yin et~al.(2023)Yin, Wu, Liang, Shi, Li, Ming, and Duan]{yin2023dragnuwa}
Shengming Yin, Chenfei Wu, Jian Liang, Jie Shi, Houqiang Li, Gong Ming, and Nan Duan.
\newblock Dragnuwa: Fine-grained control in video generation by integrating text, image, and trajectory.
\newblock \emph{arXiv: Computing Research Repo.}, abs/2308.08089, 2023.

\bibitem[Zhang et~al.(2023)Zhang, Rao, and Agrawala]{ControlNet}
Lvmin Zhang, Anyi Rao, and Maneesh Agrawala.
\newblock Adding conditional control to text-to-image diffusion models.
\newblock In \emph{Int. Conf. Comput. Vis.}, 2023.

\end{thebibliography}
}

\clearpage
\appendix
\renewcommand\thesection{\Alph{section}}
\renewcommand\thefigure{S\arabic{figure}}
\renewcommand\thetable{S\arabic{table}}
\renewcommand\theequation{S\arabic{equation}}
\setcounter{figure}{0}
\setcounter{table}{0}
\setcounter{equation}{0}

\section*{Appendix}

\section{More Implementation Details}
\label{subsec:app_implemtation}

During training, we sample 14 consecutive frames from videos, with a spatial resolution of 512$\times$320. 
Specifically, we center-crop the video to an aspect ratio of $512/320$, then resize the video frames to the resolution of $512\times320$.  
Random horizontal flip is utilized for data augmentation.
We sample the video in temporal dimension, with a frame interval of 2.
For the training of the point trajectory-based ControlNet, we sample 1 to 10 trajectories with larger motions for training. 
Specifically, we follow ReVideo~\citep{mou2024revideo} and sample the trajectories by setting the normalized lengths of the trajectories as sampling probabilities.
During ``autopilot'' mode sampling, we use the Euler sampler with 30 diffusion steps in total.
For point tracking in \cref{subsec:method_track}, we use the output feature of the second decoder block in the 3D-UNet.
We resize the shorter side of the video to the length of 512, then center crop the video to the resolution of $512\times320$.

\section{More Detailed Ablation Results}
\label{subsec:app_abltions}

\paragraph{Qualitative results for ablation studies.}
In \cref{fig:results_ablations}, we show the qualitative results for ablation studies. We supplement these results with the quantitative results in \cref{tab:ablations_7} and \cref{tab:ablations_1}, which show a similar trend to the qualitative ablation experiments.

\begin{table*}[h]
\small
\renewcommand\tabcolsep{5.0pt}
\centering
\resizebox{\textwidth}{!}{
\begin{tabular}[t]{rccccc|lcccc}
&  \multicolumn{5}{c}{\textbf{DAVIS-7}}{\hskip 0.5cm} & \multicolumn{5}{c}{\textbf{UCF101-7}}{\hskip 0.25cm}  \\
\cline{2-11}
  & PSNR$\uparrow$ & SSIM$\uparrow$ & LPIPS$\downarrow$ & FID$\downarrow$ & FVD$\downarrow$ & PSNR$\uparrow$ & SSIM$\uparrow$ & LPIPS$\downarrow$ & FID$\downarrow$ & FVD$\downarrow$ \\ 
\hline
w/o trajectory            & 20.19 & 0.6831 & 0.2787 & 28.25 & 128.71 & 24.16 & 0.8677 & 0.1798 & 32.64 & 195.54  \\
w/o traj. updating        & 20.82 & 0.7054 & 0.2621 & 27.33 & 120.73 & 24.69 & 0.8748 & 0.1842 & 31.95 & 187.37  \\
w/o bi-directional      & 20.94 & 0.7102 & 0.2602 & 27.23 & 116.81 & 24.73 & 0.8746 & 0.1845 & \textbf{31.66} & 183.74  \\
\hline
\textbf{\method (Ours)}   & \textbf{21.23} & \textbf{0.7218} & \textbf{0.2525} & \textbf{27.13} & \textbf{115.65} & \textbf{25.04} & \textbf{0.8806} & \textbf{0.1714} & 31.69 & \textbf{181.55}  \\
\hline
\end{tabular}
}
\caption{
    Ablations on each component, evaluating all 7 generated frames.
    ``w/o trajectory" denotes inference without guidance from point trajectory, ``w/o traj. updating" indicates inference without trajectory updating, and ``w/o bi" suggests trajectory updating without bi-directional consistency verification.
}
\label{tab:ablations_7}
\end{table*}

\begin{table*}[h]
\small
\renewcommand\tabcolsep{5.0pt}
\centering
\resizebox{\textwidth}{!}{
\begin{tabular}{r cccc@{\hskip 0.1cm}|cccc}
&  \multicolumn{4}{c}{\textbf{DAVIS-7 (mid-frame)}}{\hskip 0.1cm} & \multicolumn{4}{c}{\textbf{UCF101-7 (mid-frame)}}  \\
\cline{2-9}
  & PSNR$\uparrow$ & SSIM$\uparrow$ & LPIPS$\downarrow$ & FID$\downarrow$ & PSNR$\uparrow$ & SSIM$\uparrow$ & LPIPS$\downarrow$ & FID$\downarrow$ \\ 
\hline
w/o trajectory                & 19.30 & 0.6504 & 0.3093 & 57.10 & 23.14 & 0.8523 & 0.1967 & 54.98  \\
w/o traj. updating                & 19.84 & 0.6700 & 0.2935 & 55.37 & 23.60 & 0.8590 & 0.2009 & 53.83  \\
w/o bi-directional          & 19.95 & 0.6739 & 0.2919 & 54.75 & 23.65 & 0.8586 & 0.2016 & 53.54  \\
\hline
\textbf{\method (Ours)}       & \textbf{20.18} & \textbf{0.6850} & \textbf{0.2845} & \textbf{55.13} & \textbf{23.92} & \textbf{0.8646} & \textbf{0.1889} & \textbf{53.3}3  \\ 
\hline
\end{tabular}
}
\caption{
    Ablations on each component, evaluating only the middle frame out of all 7 generated frames.
    ``w/o trajectory" denotes inference without guidance from point trajectory, ``w/o traj. updating" indicates inference without trajectory updating, and ``w/o bi" suggests trajectory updating without bi-directional consistency verification.
}
\label{tab:ablations_1}
\end{table*}

\begin{figure*}
    \centering
    \includegraphics[width=\textwidth]{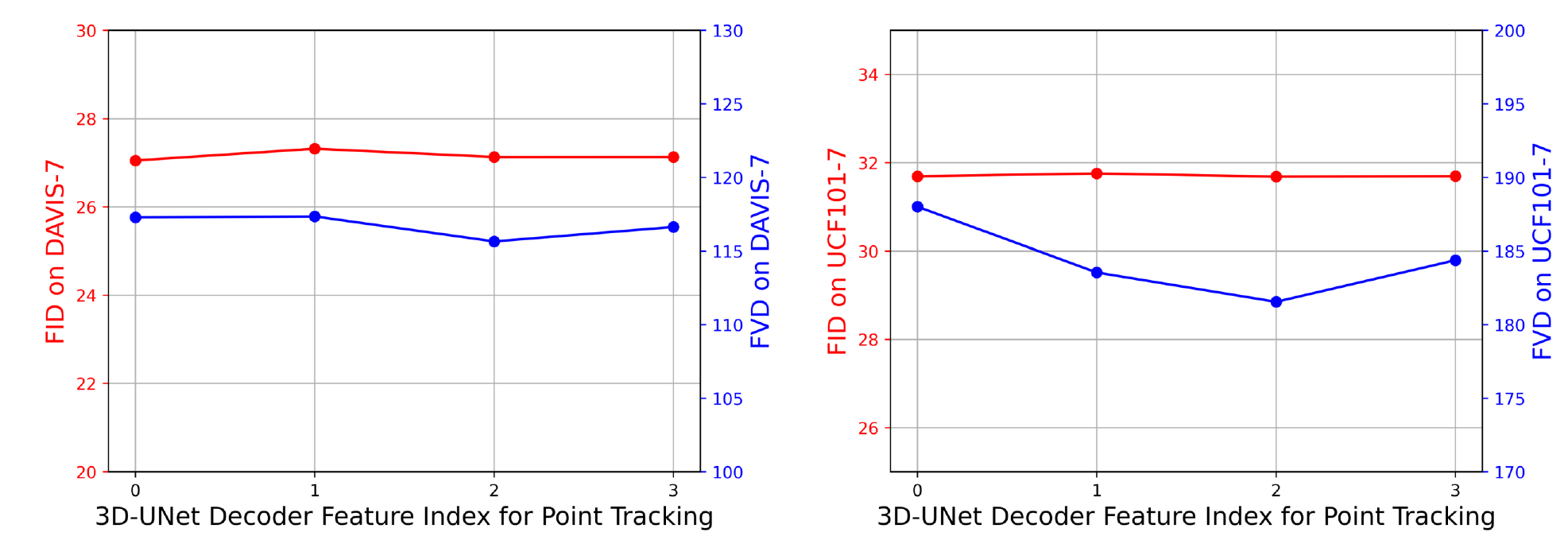}
    \caption{%
        Ablations on diffusion feature for point tracking at test time, experiments conducted on DAVIS-7 (left) and UCF101-7 (right).
    }
    \label{fig:app_ablations_featidx}
\end{figure*}

\paragraph{Ablations on diffusion feature for point tracking.}
As detailed in \cref{subsec:method_track}, we perform point tracking using the diffusion feature for point trajectory updating. Here we perform ablated experiments on the selection of the diffusion feature. The results are shown in \cref{fig:app_ablations_featidx}. It can be seen that in both DAVIS-7 and UCF-7, point tracking with the output feature from the second diffusion block gives rise to the best-performing results in FVD.

\begin{figure*}
    \centering
    \includegraphics[width=\textwidth]{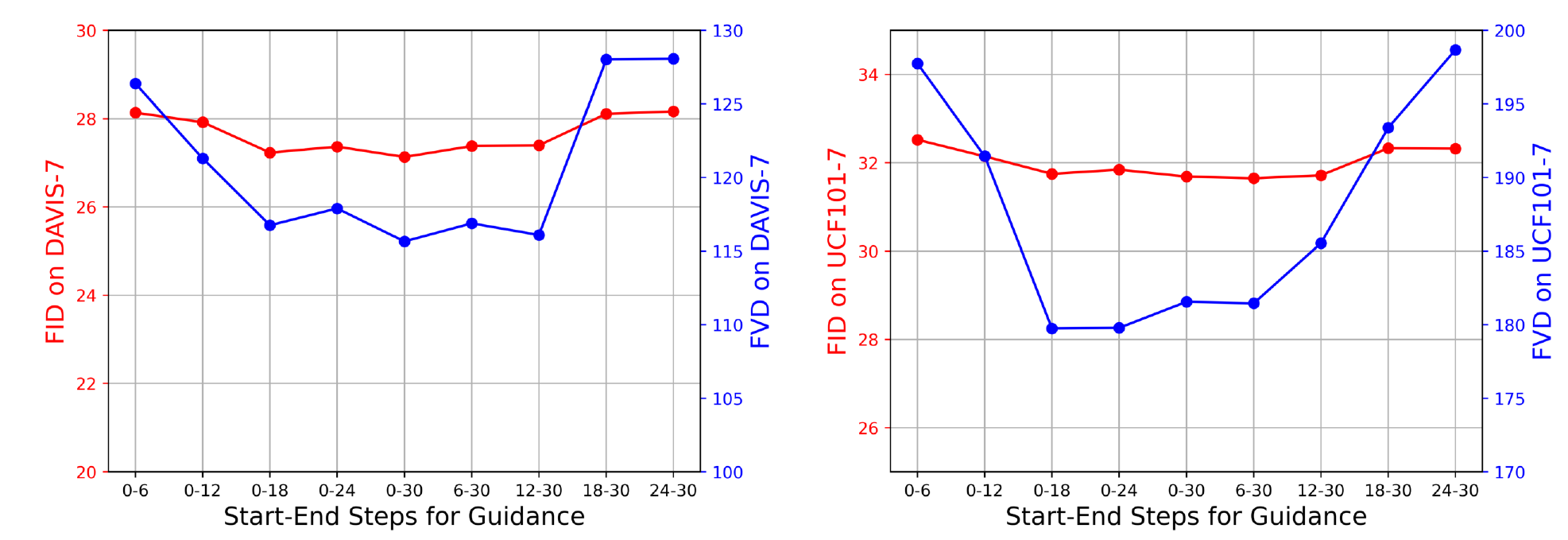}
    \caption{%
        Ablations on the start and end diffusion steps for correspondence guidance, experiments conducted on DAVIS-7 (left) and UCF101-7 (right). We use a total sampling step of 30.
    }
    \label{fig:app_ablations_step}
\end{figure*}

\paragraph{Ablations on diffusion steps for correspondence guidance.}
We ablate the diffusion steps for correspondence guidance by only applying the guidance at the early steps or late steps in diffusion sampling. The results are shown in \cref{fig:app_ablations_step}.
As can be seen, the early steps are often more important than the late steps for correspondence modeling. 
For example, on DAVIS-7, a pleasing FVD can be obtained when performing guidance only on 0-18 diffusion steps. By contrast, performing guidance only on 18-30 diffusion steps brings litter improvements.
We speculate that this is because the early diffusion steps focus on the structural information of the video, while the late diffusion steps focus on the texture and details~\citep{xue2024raphael}.
%
The correspondence guidance at early steps already helps the model obtain a reasonable video structure.
In the implementation, we simply apply correspondence guidance in all diffusion steps, without detailed searches on the hyper-parameter.

\begin{figure*}
    \centering
    \includegraphics[width=\textwidth]{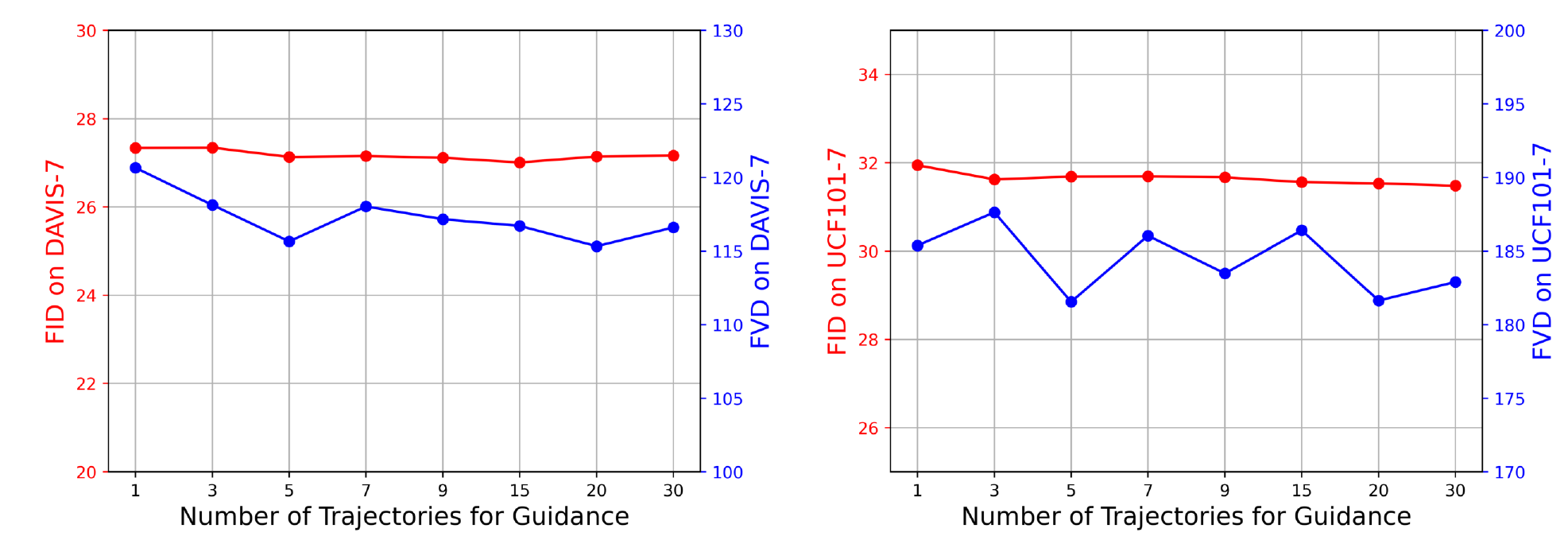}
    \caption{%
        Ablations on the number of trajectories for guidance during sampling, experiments conducted on DAVIS-7 (left) and UCF101-7 (right).
    }
    \label{fig:app_ablations_tracks_n}
\end{figure*}

\paragraph{Ablations on the number of trajectories for correspondence guidance.}
As described in \cref{subsec:method_track}, we use $m$ trajectories for correspondence guidance during sampling. 
Here we perform ablated experiments on this hyper-parameter, and the result is shown in \cref{fig:app_ablations_tracks_n}.
It can be seen that sampling with the  5 trajectories leads to the best performance. Thus we set $m=5$ by default.

\section{More Details on Comparison with Previous Methods}
\paragraph{Benchmark.}
We follow the practice of VIDIM~\citep{jain2024video} and perform the quantitative evaluation on DAVIS-7 and UCF101-7 datasets using both reconstruction and generative metrics.
Both DAVIS-7 and UCF101-7 are obtained by sampling 7 consecutive video frames from the corresponding datasets. 
We use all videos in the DAVIS dataset and a subset of 400 videos in the UCF101 dataset.

\begin{figure*}
    \centering
    \includegraphics[width=\textwidth]{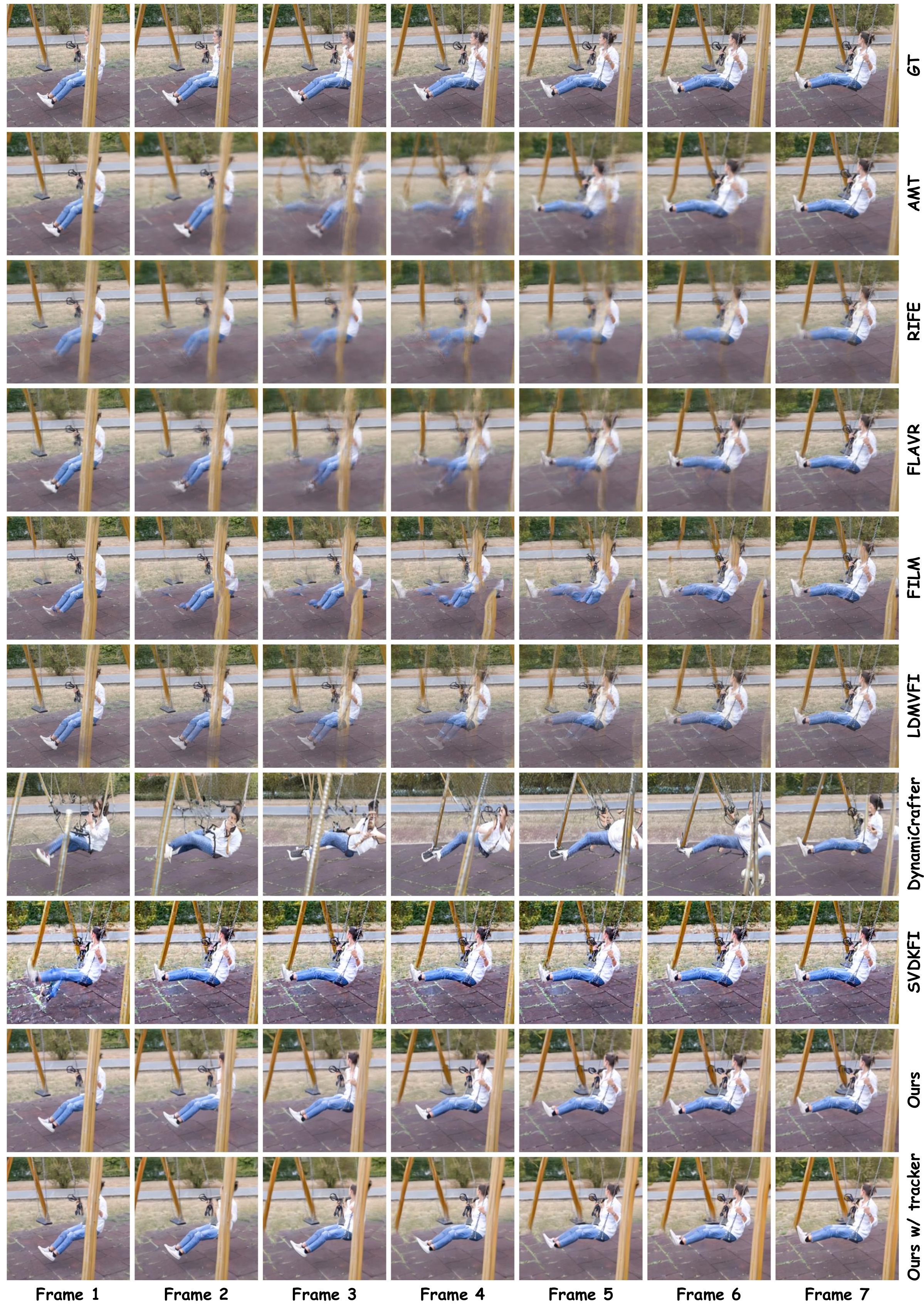}
    \caption{%
        More qualitative comparison with existing methods. 
        ``GT" strands for ground truth.
    }
    \label{fig:app_comparison_1}
\end{figure*}

\begin{figure*}
    \centering
    \includegraphics[width=\textwidth]{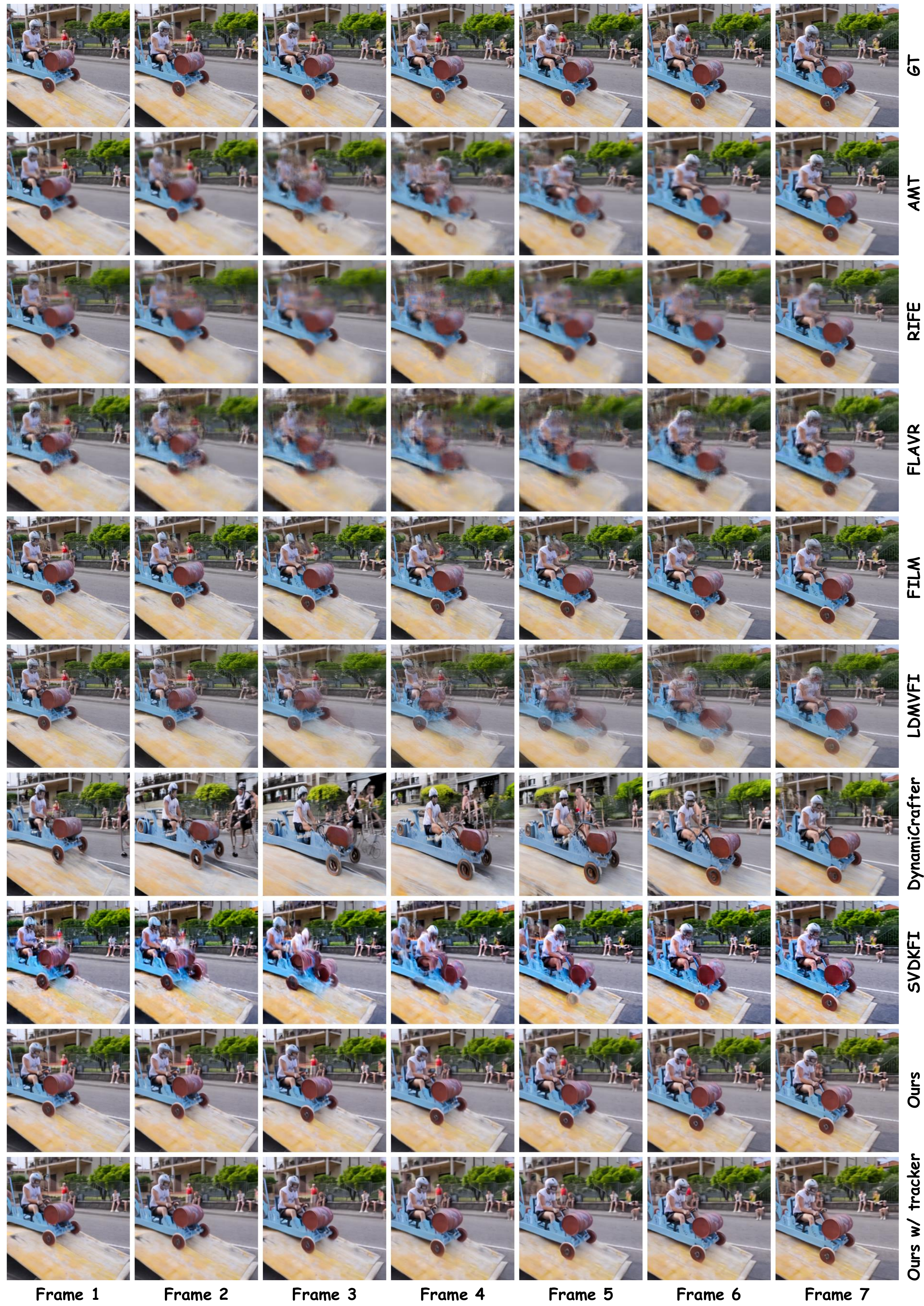}
    \caption{%
        More qualitative comparison with existing methods. 
        ``GT" strands for ground truth.
    }
    \label{fig:app_comparison_2}
\end{figure*}

\begin{figure*}
    \centering
    \includegraphics[width=\textwidth]{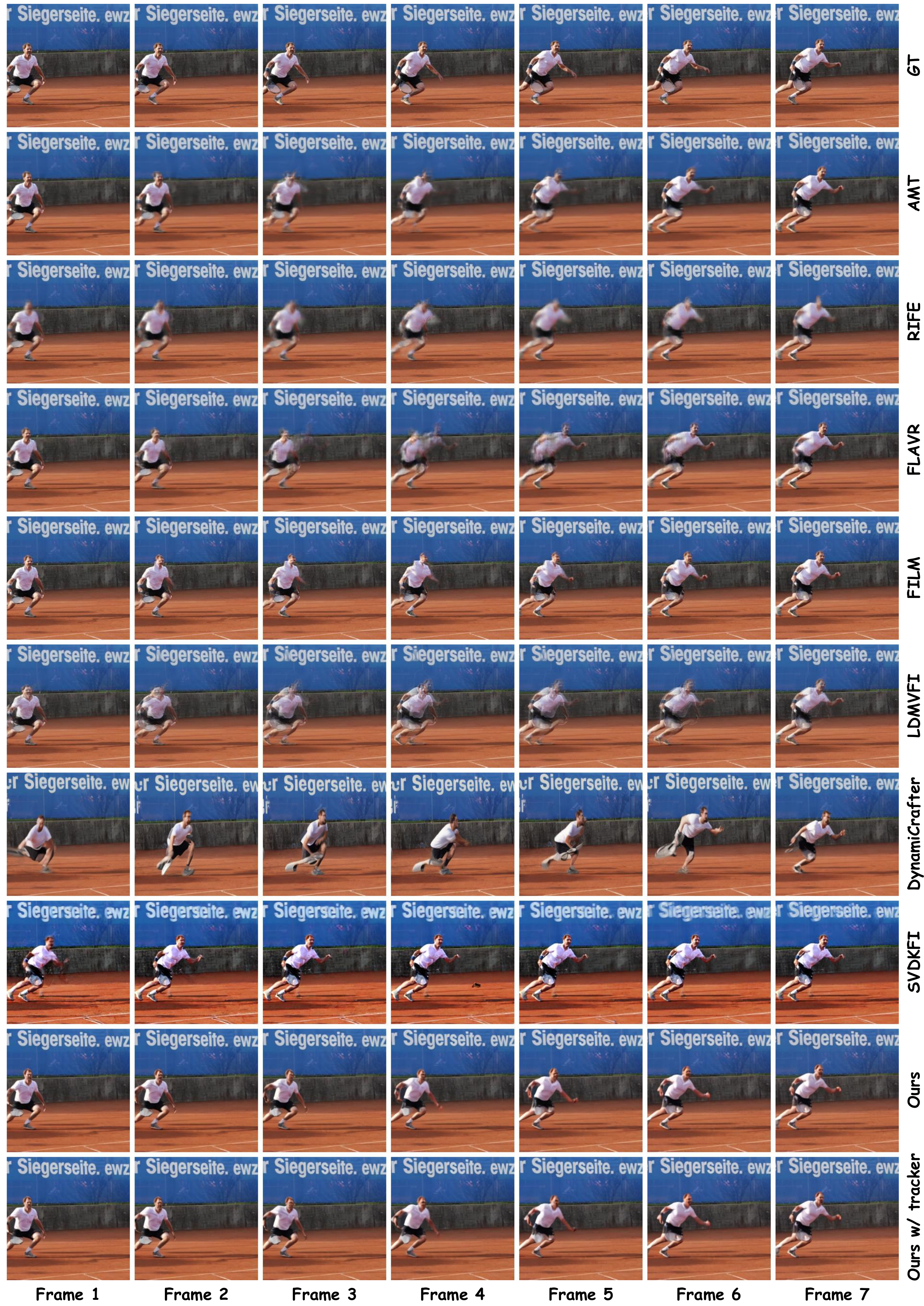}
    \caption{%
        More qualitative comparison with existing methods. 
        ``GT" strands for ground truth.
    }
    \label{fig:app_comparison_3}
\end{figure*}

\begin{figure*}
    \centering
    \includegraphics[width=\textwidth]{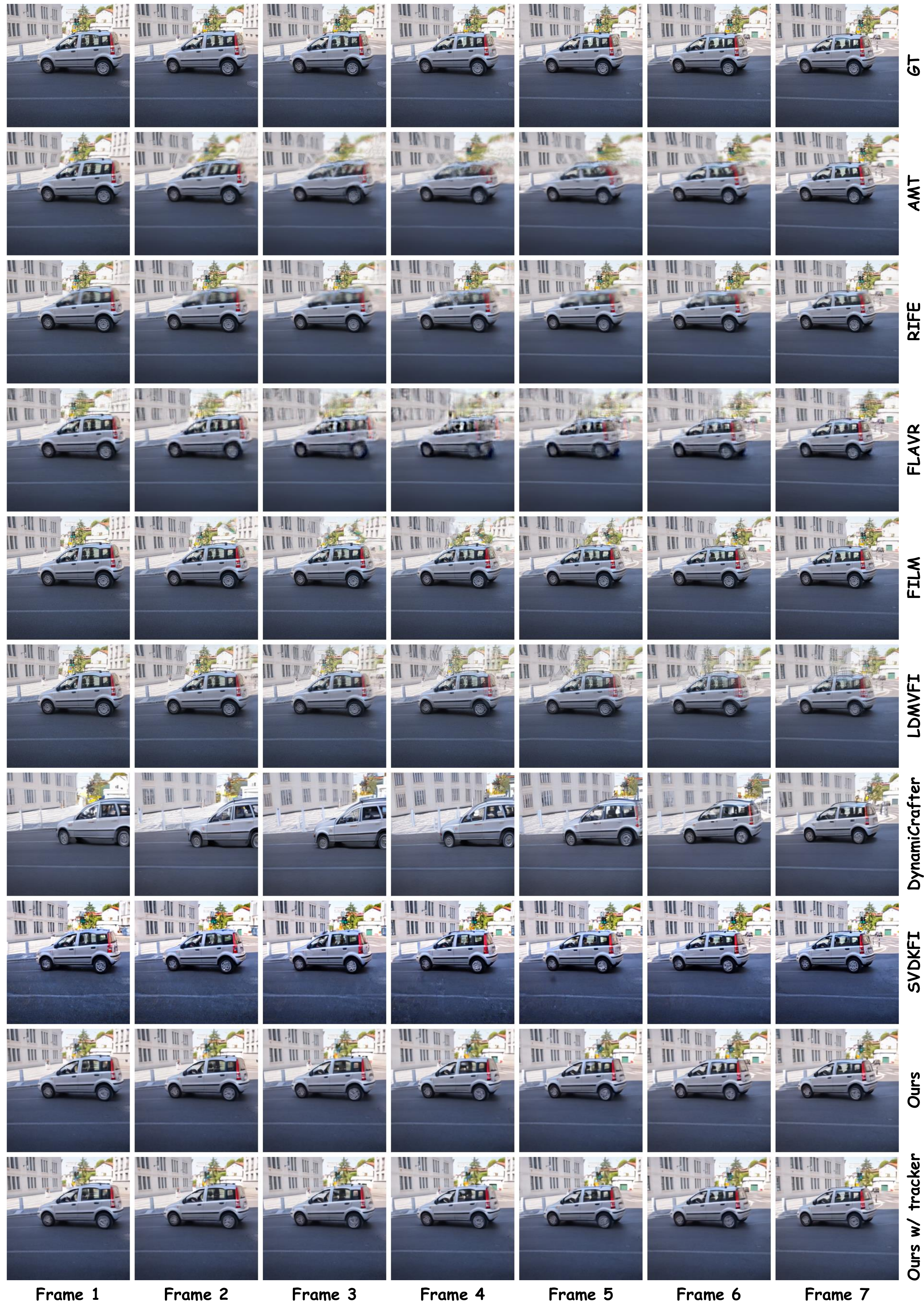}
    \caption{%
        More qualitative comparison with existing methods. 
        ``GT" strands for ground truth.
    }
    \label{fig:app_comparison_4}
\end{figure*}

\paragraph{More results on comparisons.}
In \cref{tab:compare_1} we provide the quantitative comparison based on the middle frame of the 7 interpolated video frames.
Besides, in \cref{fig:app_comparison_1}, \cref{fig:app_comparison_2}, \cref{fig:app_comparison_3}, and \cref{fig:app_comparison_4}, we show more qualitatively comparisons with exiting methods.

\begin{table*}[t]
\small
\renewcommand\tabcolsep{5.0pt}
\centering
\resizebox{\textwidth}{!}{
\begin{tabular}{r cccc@{\hskip 0.1cm}|cccc}
&  \multicolumn{4}{c}{\textbf{DAVIS-7 (mid-frame)}}{\hskip 0.1cm} & \multicolumn{4}{c}{\textbf{UCF101-7 (mid-frame)}}  \\
\cline{2-9}
  & PSNR$\uparrow$ & SSIM$\uparrow$ & LPIPS$\downarrow$ & FID$\downarrow$ & PSNR$\uparrow$ & SSIM$\uparrow$ & LPIPS$\downarrow$ & FID$\downarrow$ \\ 
\hline
AMT \citep{licvpr23amt}                         & 20.59 & 0.6834 & 0.3564 & 100.36 & 25.24 & 0.8837 & 0.2237 & 75.97  \\ 
RIFE \citep{huang2020rife}                      & \textbf{20.74} & 0.6813 & 0.3102 & 80.78  & \textbf{25.68} & \textbf{0.8842} & 0.1835 & 59.33  \\ 
FLAVR \citep{flavr}                              & 19.93 & 0.6514 & 0.4074 & 118.45 & 24.93 & 0.8796 & 0.2164 & 79.86  \\ 
FILM \citep{reda2022film}                       & 20.28 & 0.6671 & \textbf{0.2620} & \textbf{48.70}  & 25.31 & 0.8818 & \textbf{0.1623} & \textbf{41.23}  \\ 
LDMVFI \citep{danier2024ldmvfi}                 & 19.87 & 0.6435 & 0.2985 & 56.46  & 25.16 & 0.8789 & 0.1695 & 43.01  \\ 
DynamicCrafter \citep{xing2023dynamicrafter}    & 14.61 & 0.4280 & 0.5082 & 77.65 & 17.05  & 0.6935 & 0.3502 & 97.01  \\ 
SVDKFI \citep{wang2024generative}                                 & 16.06 & 0.4974 & 0.3719 & 53.49 & 20.03 & 0.7775 & 0.2326 & 69.26  \\ 
\hline
\textbf{\method (Ours)}                           & 20.18 & \textbf{0.6850} & 0.2845 & 55.13 & 23.92 & 0.8646 & 0.1889 & 53.33  \\ 
\demphs{\method with Co-Tracker (Ours)}  & \demphs{21.94} & \demphs{0.7693} & \demphs{0.2437} & \demphs{55.77} & \demphs{25.86} & \demphs{0.8868} & \demphs{0.1873} & \demphs{54.64}  \\ 
\end{tabular}
}
\caption{
Quantitative comparison with existing video interpolation methods on reconstruction and generative metrics, evaluated only on the middle frame out of all 7 generated frames.
}
\label{tab:compare_1}
\end{table*}

\section{More Qualitative Results}

\begin{figure*}
    \centering
    \includegraphics[width=\textwidth]{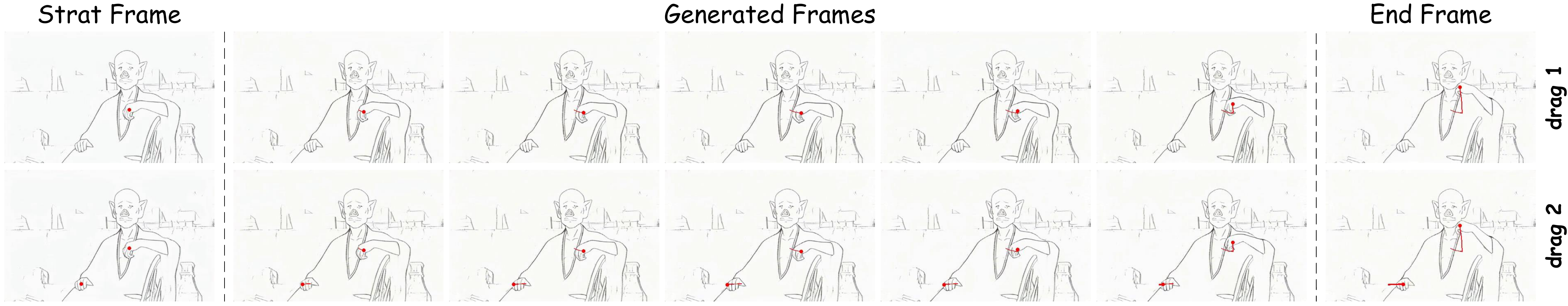}
    \caption{%
    {More results on user interaction.} 
    We show the results of two trajectory controls with the same input image pair.
    }
    \label{fig:app_drag}
\end{figure*}

\begin{figure*}
    \centering
    \includegraphics[width=\textwidth]{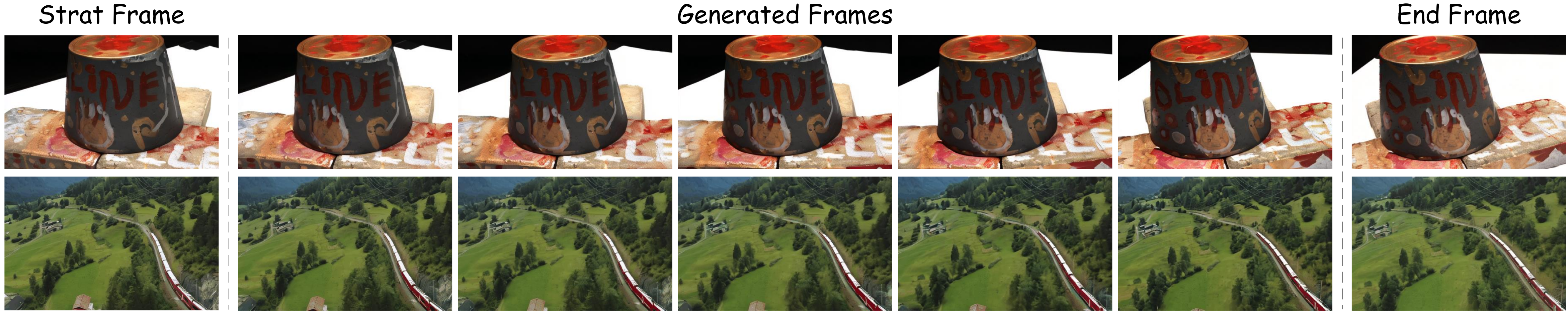}
    \caption{%
        {More results on novel view synthesis.} The first and second rows show results on static and dynamic scenes, respectively. 
    }
    \label{fig:app_nvs}
\end{figure*}

\begin{figure*}
    \centering
    \includegraphics[width=\textwidth]{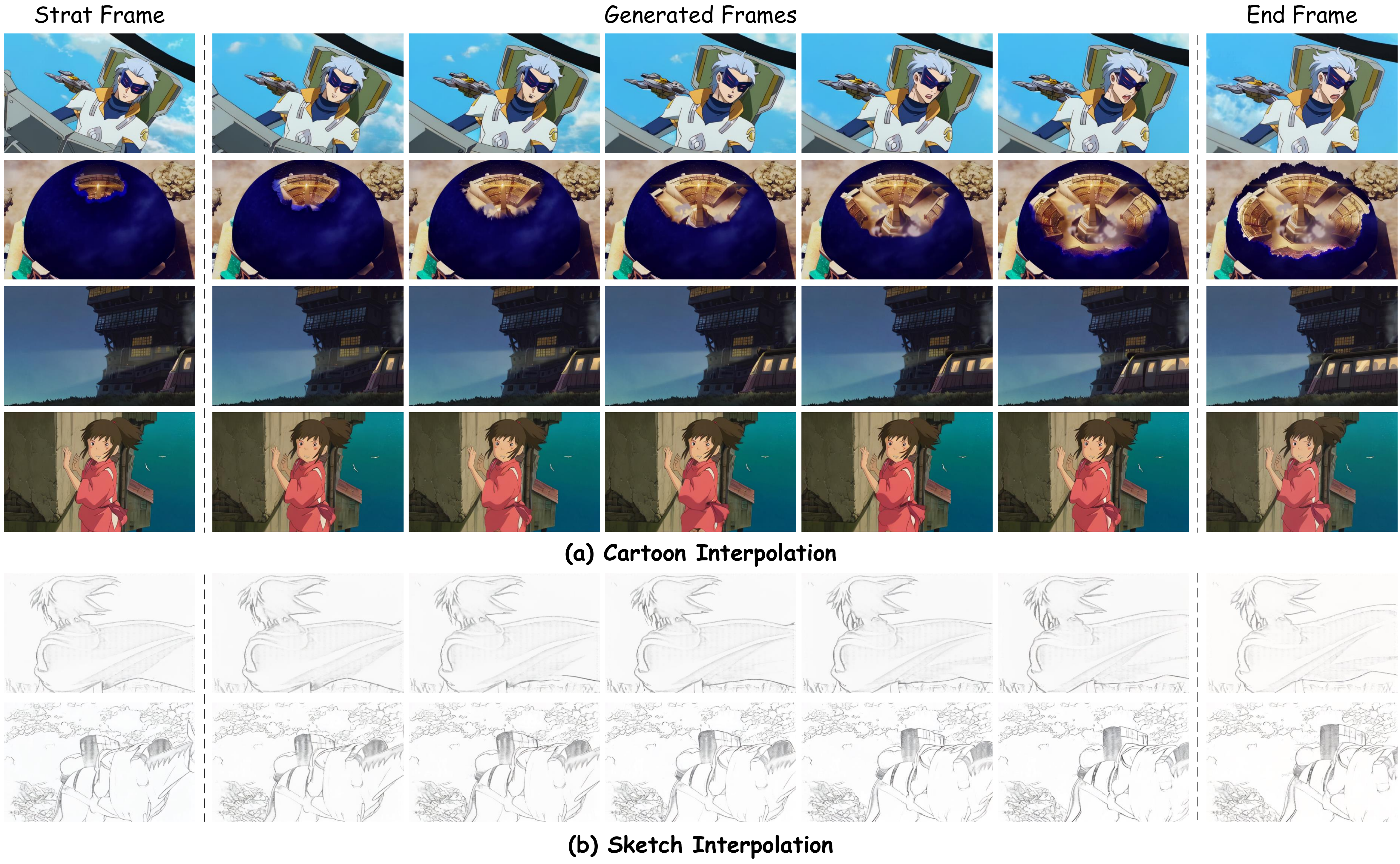}
    \caption{%
        More results on (a) cartoon and (b) sketch interpolation.
    }
    \label{fig:app_toon}
\end{figure*}

\begin{figure*}
    \centering
    \includegraphics[width=\textwidth]{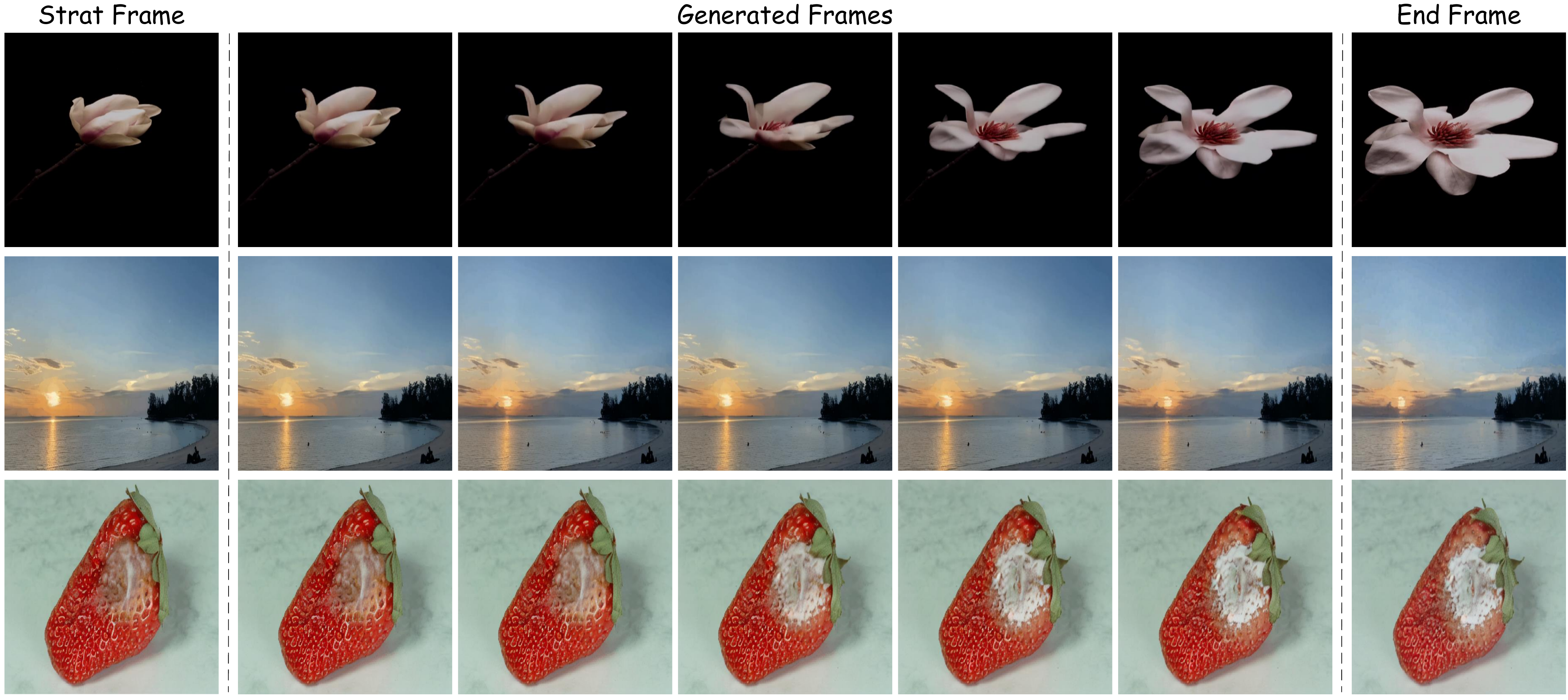}
    \caption{%
        More results on time-lapsing video generation.
    }
    \label{fig:app_chron}
\end{figure*}

\begin{figure*}
    \centering
    \includegraphics[width=\textwidth]{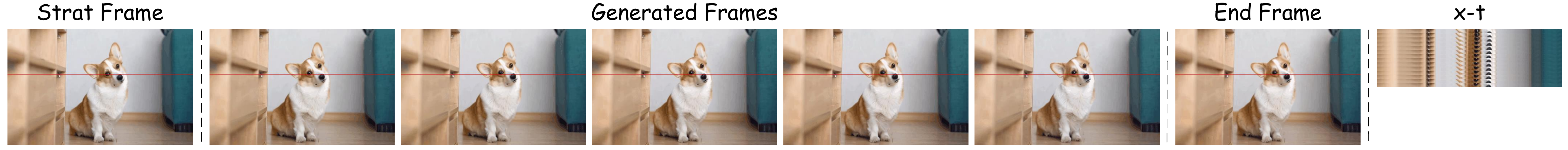}
    \caption{%
        More results on slow-motion video generation. The x-t slice highlighted in red on video frames is visualized on the right.
    }
    \label{fig:app_slow}
\end{figure*}

\begin{figure*}
    \centering
    \includegraphics[width=\textwidth]{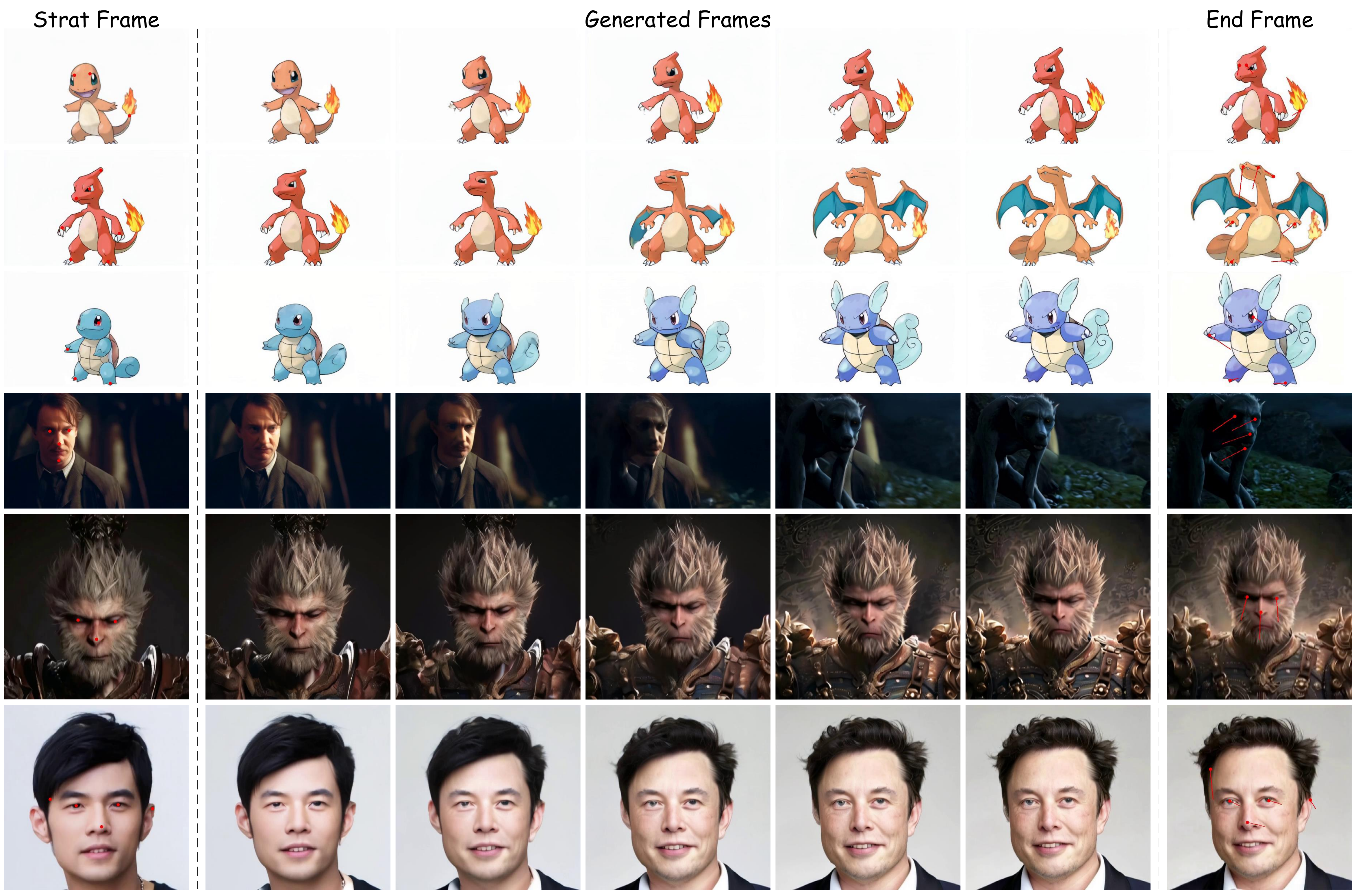}
    \caption{%
    More results on image morphing.
    }
    \label{fig:app_morph}
\end{figure*}

We provide more qualitative results on drag control, novel view synthesis, cartoon and sketch interpolation, time-lapsing video generation, slow-motion video generation, and image morphing in \cref{fig:app_drag}, \cref{fig:app_nvs}, \cref{fig:app_toon}, \cref{fig:app_chron}, \cref{fig:app_slow}, and \cref{fig:app_morph}, respectively.

\section{Discussions on Limitations}

\method is built on top of the large-scale pre-trained video diffusion model, thus it inherits the limitations of the pre-trained model.
Moreover, the point trajectories in \method rely on the matching points between the input image pair for interpolating complex motions. 
While this is a step forward compared with current models that can only simply motions, our method still faces difficulties when the differences between the front and back frames are so large that no matched points can be found at all.
Thus, we will explore more powerful pre-trained video diffusion models, as well as training video interpolation models on larger-scale video data in the future.
Lastly, our approach currently only supports drag control and does not explore other interaction methods. 
In the future, we will continue to explore other user-friendly controls such as text control and camera pose control.

\end{document}